%% file: sn-article.tex
\documentclass[pdflatex,sn-nature]{sn-jnl}% Style for submissions to Nature Portfolio journals
\usepackage{graphicx}%
\usepackage{multirow}%
\usepackage{amsmath,amssymb,amsfonts}%
\usepackage{amsthm}%
\usepackage{mathrsfs}%
\usepackage[title]{appendix}%
\usepackage{xcolor}%
\usepackage{textcomp}%
\usepackage{manyfoot}%
\usepackage{booktabs}%
\usepackage{algorithm}%
\usepackage{algorithmicx}%
\usepackage{algpseudocode}%
\usepackage{listings}%
\usepackage[table]{xcolor}
\usepackage{booktabs,tabularx,array}
\newcolumntype{L}{>{\raggedright\arraybackslash}X}
\usepackage{subcaption}

\theoremstyle{thmstyleone}%
\theoremstyle{thmstyletwo}%

\theoremstyle{thmstylethree}%

\begin{document}

\title[The Potential and Limitations of Vision-Language Models for Human Motion Understanding: A Case Study in Data-Driven Stroke Rehabilitation]{
The Potential and Limitations of Vision-Language Models for Human Motion Understanding: A Case Study in Data-Driven Stroke Rehabilitation}


\author[1]{\fnm{Victor} \sur{Li}}
\author[1,2]{\fnm{Naveenraj} \sur{Kamalakannan}}
\author[3]{\fnm{Avinash} \sur{Parnandi}}
\author[4,5]{\fnm{Heidi} \sur{Schambra}}\email{Heidi.Schambra@nyulangone.org}\equalcont{Equal contribution.}
\author[1,6]{\fnm{Carlos} \sur{Fernandez-Granda}}\email{cfgranda@cims.nyu.edu}\equalcont{Equal contribution.}

\affil*[1]{\orgdiv{Center for Data Science}, \orgname{New York University}, \orgaddress{\street{60 Fifth Ave}, \postcode{10011}, \city{New York}, \state{NY}, \country{USA}}}

\affil*[2]{\orgdiv{Tandon School of Engineering}, \orgname{New York University}, \orgaddress{\street{6 MetroTech Center}, \postcode{11201}, \city{Brooklyn}, \state{NY}, \country{USA}}}

\affil*[3]{\orgname{VitalConnect}, \orgaddress{\street{2870 Zanker Road \#100}, \postcode{95134}, \city{San Jose}, \state{CA}, \country{USA}}}

\affil*[4]{\orgdiv{Department of Neurology}, \orgname{NYU Grossman School of Medicine}, \orgaddress{\street{550 1st Ave}, \postcode{10016}, \city{New York}, \state{NY}, \country{USA}}}

\affil*[5]{\orgdiv{Department of Rehabilitation Medicine}, \orgname{NYU Grossman School of Medicine}, \orgaddress{\street{550 1st Ave}, \postcode{10016}, \city{New York}, \state{NY}, \country{USA}}}

\affil*[6]{\orgdiv{Courant Institute of Mathematical Sciences}, \orgname{New York University}, \orgaddress{\street{251 Mercer St}, \postcode{10012}, \city{New York}, \state{NY}, \country{USA}}}

\abstract{Vision–language models (VLMs) have demonstrated remarkable performance across a wide range of computer-vision tasks, sparking interest in their potential for digital health applications. Here, we apply VLMs to two fundamental challenges in data-driven stroke rehabilitation: automatic quantification of rehabilitation dose and impairment from videos. We formulate these problems as motion-identification tasks, which can be addressed using VLMs. We evaluate our proposed framework on a cohort of 29 healthy controls and 51 stroke survivors. Our results show that current VLMs lack the fine-grained motion understanding required for precise quantification: dose estimates are comparable to a baseline that excludes visual information, and impairment scores cannot be reliably predicted. Nevertheless, several findings suggest future promise. With optimized prompting and post-processing, VLMs can classify high-level activities from a few frames, detect motion and grasp with moderate accuracy, and approximate dose counts within 25\% of ground truth for mildly impaired and healthy participants, all without task-specific training or finetuning. These results highlight both the current limitations and emerging opportunities of VLMs for data-driven stroke rehabilitation and broader clinical video analysis.}


\maketitle

\input{sections/01_intro}

\input{sections/02_results}
\input{sections/03_discussion}
\input{sections/04_method}

\backmatter

%\section*{Declarations}
% \subsection*{Funding}
\subsection*{Acknowledgements}

V.L., N.K., H.S. and C.F.G. were supported by NSF grant 2404476.  

% \subsection*{Competing interests}

\subsection*{Ethics approval and consent to participate}

For the StrokeRehab dataset, all subjects provided written informed consent in accordance with the Declaration of Helsinki. The study was approved by the Institutional Review Board at the New York University Grossman School of Medicine.

% \subsection*{Consent for publication}

% \subsection*{Data Availability}

% \noindent
% If any of the sections are not relevant to your manuscript, please include the heading and write `Not applicable' for that section. 

\bigskip

% Springer journals and proceedings: \url{https://www.springer.com/gp/editorial-policies}

% Nature Portfolio journals: \url{https://www.nature.com/nature-research/editorial-policies}

% \textit{Scientific Reports}: \url{https://www.nature.com/srep/journal-policies/editorial-policies}

% BMC journals: \url{https://www.biomedcentral.com/getpublished/editorial-policies}
% \end{flushleft}

\bibliography{sn-bibliography}% common bib file

%%%%%%%%%%%%%%%%%%%%%%%% APPENDIX %%%%%%%%%%%%%%%%%%%%%%
\appendix
\input{sections/05_appendix}


\end{document}

%% file: sections/01_intro.tex
\section*{Introduction}\label{sec:intro}

%%%%%%%%%%%%%%%%%%%%%%%%

\begin{figure}
    \centering
    \includegraphics[width=\linewidth]{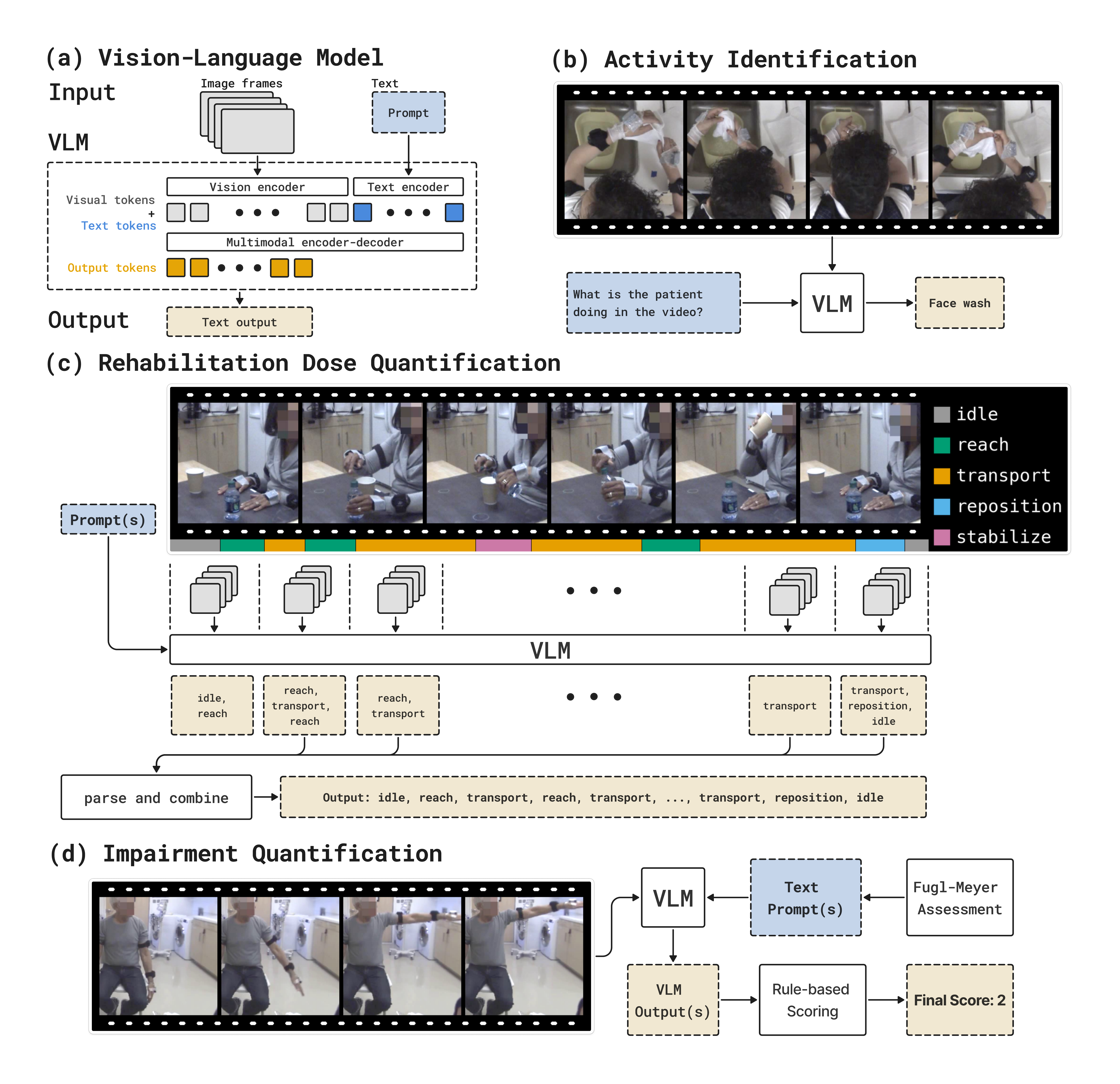}
    \vspace{0.3cm}
    \caption{\textbf{Vision-Language Models (VLMs) for Data-Driven Stroke Rehabilitation.} \textbf{(a)} A VLM can function as a video question-answering system. The question, or \textit{prompt}, and image frames from the video are separately encoded into tokens, which are fed-forward through a transformer-based backbone. The backbone's output is then decoded into text. \textbf{(b)} \emph{Activity identification:} The VLM is provided $8$ frames uniformly sampled from a video ($4$ frames are shown due to space constraints) and a description of nine rehabilitation activities. 
    The VLM output classifies the activity in the video. \textbf{(c)} \emph{Dose quantification:} The VLM is provided a video segment along with a textual prompt. Its output is then utilized to classify the segment into one of five functional motions or primitives. Rehabilitation dose is quantified by counting the primitives over the whole video. 
    \textbf{(d)} \emph{Impairment quantification:} The VLM processes video segments of subjects performing mobility exercises from the Fugl-Meyer Assessment (FMA), a standard clinical evaluation of impairment. The input prompt is the corresponding FMA item. The outputs are aggregated to estimate the FMA score of the subject.
}
    \label{fig:fig1}
\end{figure}

Vision-language models (VLMs) are large-scale deep learning models trained on massive multimodal datasets to jointly interpret visual and textual information. These models have recently achieved remarkable performance across a wide range of computer vision tasks, including image and video captioning, visual question answering, optical character recognition, document understanding, and open-vocabulary object detection \citep{zhang2024llavanextvideo,li2024llava,liu2024nvila,bai2025qwen25vltechnicalreport,zhu2025internvl3exploringadvancedtraining,wang2025internvl3,lin2023video,chen2024expanding,zhang2024long,yang2025kwaikeyevl15technical,comanici2025gemini,openai2023gpt4,anthropic2024claude3_5}. In this work, we investigate the use of VLMs for analyzing human motion, with a particular emphasis on stroke rehabilitation. 

Stroke is the leading cause of disability worldwide. In 2021, there were 93.8 million stroke survivors in the world \cite{GBD2021Stroke}, with upper limb impairment occurring in approximately 50–80\% of cases~\citep{2023_stroke_recs}. Most patients remain unable to perform daily activities independently six months post-stroke, reducing quality of life and imposing enormous societal and economic burdens. In animal models of stroke, high numbers of movement repetitions have been shown to improve abnormalities in motion quality ~\citep{motor_skill_training_rats,Murata2008,jeffers_stroke_rehab}. Unfortunately, clinical rehabilitation research has largely failed to translate this fundamental work from animals to humans~\cite{Saikaley2023,Stinear2020,CumberlandConsensusWorkingGroup2009}. A major barrier is the absence of precise, practical methods for quantifying both training dose (i.e., the number of movements performed during rehabilitation) and impairment level (i.e., the quality of those movements)~\cite{Lohse2018,Walker2017}. 

Here, we evaluate the potential of VLMs to automatically quantify rehabilitation dose and assess impairment from video data. Both of these tasks can be formulated as human-motion recognition problems, where the goal is to identify subtle movements. VLMs hold great promise for these and other digital health tasks involving human motion, such as remote patient monitoring, interactive diagnostic assistance, gait analysis, and surgical training. 
VLMs receive freeform textual input, called a ``prompt," which can be flexibly used to determine what information should be extracted from the visual input. This flexibility makes it possible to make clinically-informed queries to the models.  

Previous approaches utilizing machine learning for automatic dose quantification \cite{kaku2020towards,kaku2022strokerehab,parnandi2022primseq} and impairment quantification \cite{yu2024quantifying,Parnandi2023_MovementAbnormalityStroke} from video and wearable-sensor data relied on training datasets with similar characteristics to the test data. This reliance on similar data represents a fundamental limitation for such approaches, given that training data in this domain is very scarce (the largest publicly available dataset contains only 51 subjects \cite{kaku2022strokerehab}). In contrast, VLMs are, in principle, able to directly process images with very different lighting, backgrounds, camera angles and content, as they are trained on enormous amounts of data. In that sense, these models could represent a transformative paradigm shift in digital health applications, where video data are highly heterogeneous. 

% Stroke is very important because XXX. Currently rehabilitation is not very data-driven, which is problematic. In order to achieve data-driven stroke rehabilitation we need to quantify dose and impairment.

% Here we explore how to address these two tasks from video data. Doing these kind of tasks automatically from videos would be very useful. Telehealth, monitoring, enable clinical trials. 

% VLMs are well suited for this because both tasks can be reformulated in terms of specific questions that can be encoded as prompts, as shown in Figure 1. For dose quantification... For impairment quantification... A compelling property of VLMs is that they are in principle able to process very heterogeneous videos with different backgrounds, camera angles, subjects, etc. given the vast amount of data they are trained on. This is great for stroke rehabilitation, since rehabilitation sessions can be recorded under very different conditions.  

%Previous works applied more traditional ML where they train from scratch on training dataset. Very challenging because datasets are minuscule compared to those required for state-of-the-art computer vision. Cannot be expected to generalize robustly. Largest publicly-available is StrokeRehab which only has around 70 subjects (and raw videos are not available).

Figure~\ref{fig:fig1} shows how we propose to apply VLMs for dose and impairment quantification from videos, and also for identification of high-level rehabilitation activities. Specialized prompts are provided to the model, along with multiple video frames, in order to solicit the information required for each task. In the case of dose quantification, the VLM is asked to identify five basic functional movements or \emph{primitives}, which are then counted to determine the rehabilitation dose~\cite{schambra2019taxonomy}. In the case of impairment quantification, each prompt is based on an item from the Fugl-Meyer Assessment (FMA), which is a standardized 33-item clinical scale used to quantify motor impairment of the arm, wrist, and hand after stroke~\cite{fugl1975method}. The model outputs are then combined to produce an estimate of the FMA score. The Methods section provides a more detailed description of our approach.

We evaluate the ability of VLMs to quantify both training dose and motor impairment using a cohort comprising 29 healthy controls and 51 stroke survivors (Table~\ref{tab:stroke_cohort}). Our results indicate that %VLMs cannot identify fine-grained human motion with sufficient accuracy to achieve precise quantification. In the case of dose, the models achieve similar performance to a baseline that does not take into account the video data, only the transition statistics between functional primitives. In the case of impairment, the VLM is not able to reliably answer any of the items in the Fugl-Meyer Assessment. 
current VLMs lack the fine-grained understanding of human motion required for precise quantification. For dose estimation, VLM performance is comparable to a baseline that relies solely on transition statistics between functional primitives, without access to the video data. For impairment assessment, the VLM fails to consistently provide correct responses for any of the items in the Fugl–Meyer Assessment.

Despite these negative results, several of our findings are moderately encouraging and point to the future potential of VLMs in data-driven stroke rehabilitation and other digital medicine applications. We observe that these models can classify high-level activities from very few frames. %, provided that prompts are optimized on held-out subjects. 
They can also identify the presence of motion and grasp with moderate accuracy. For a highly structured rehabilitation task, incorporating carefully engineered postprocessing yields dose estimates within 25\% of the ground truth for mildly impaired and healthy subjects. These results require optimizing the prompts on a few held-out individuals, but no additional training or finetuning.%, which showcases the robustness of VLMs when processing real-world data. 

%% file: sections/02_results.tex
\section*{Results}
\label{sec:results}

\begin{table}[t]
\centering
\caption{Demographic and clinical characteristics of the cohort (left) and task-specific Held-out and Test splits used for VLM evaluation in the Results section (right). Mean $\pm$ standard deviation is reported where applicable.}
\label{tab:stroke_cohort}

% Two-column wrapper: left flex (X), right narrower fixed width (p{...})
\begin{tabularx}{\linewidth}{@{}>{\raggedright\arraybackslash}p{0.5\linewidth}@{\hspace{0.03\linewidth}}p{0.43\linewidth}@{}}

% -------- LEFT PANEL --------
{\small
\begin{tabularx}{\linewidth}{@{}l >{\raggedright\arraybackslash}X@{}}
\toprule
\textbf{Characteristic} & \textbf{Value} \\
\midrule
\# of subjects & 80 \\
\# of Trials & 3448 \\
Age & $59.2 \pm 13.7$ \\
Sex & 38 Male, 42 Female \\
Race & 34W, 26B, 9A, 1AI, 10O \\
Paretic Side & 28 Left, 23 Right, 29 n/a \\
Fugl--Meyer Score & $65.1 \pm 1.1$ (Control), $43.5 \pm 16.2$ (Stroke) \\
Impairment Level & 29 C, 20 Mi, 23 Mo, 8 S \\
Time Since Stroke & $5.4 \pm 6.1$ (Stroke) \\
\bottomrule
\end{tabularx}
}
&
% -------- RIGHT PANEL (narrower) --------
{\small
\begin{tabularx}{\linewidth}{@{}L L@{}}
\toprule
\textbf{Task} & \textbf{Impairment Level (C, Mi, Mo, S)} \\
\midrule
(A) Activity &  TC: 18, 20, 23, 8\\
 Identification & PO: 2, 0, 0, 0 \\\midrule
(B) Dose & TC: 5, 3, 2, 0 \\
 Quantification &  \\\midrule
(C) Dose Quant. & TC: 5, 5, 5, 5 \\
 RTT/Shelf & PO: 2, 0, 0, 1 \\\midrule
(D) Impairment & TC: 4, 10, 10, 4 \\
 Quantification \\%& PO: 1, 1, 1, 1 \\
\bottomrule
\end{tabularx}
}

\end{tabularx}

\vspace{4pt}
\textit{Abbreviations:} 
W = White; B = Black; A = Asian; AI = American Indian; O = Other; 
C = Control/Healthy Subjects; Stroke = Stroke Subjects of various impairment levels based on their UE Fugl-Meyer score (0-66); Mi = Mild (53-65); Mo = Moderate (26-52); S = Severe (0-25); 
TC = Test cohort; PO = Held-out subjects used for prompt optimization. Age and time since stroke are in years

\end{table}

\paragraph{Cohort and evaluation procedure.}

This study is based on a cohort of $80$ individuals presented in Table~\ref{tab:stroke_cohort}, consisting of $29$ healthy subjects and $51$ stroke patients. The stroke patients are separated into three motor impairment levels---mild, moderate, severe---based on the Fugl-Meyer assessment (FMA), a standard impairment evaluation for stroke survivors~\cite{fugl1975method}.  We apply two evaluation procedures:

\begin{itemize}
    \item \textbf{Independent evaluation:} The cohort is treated as a fully independent external test cohort. The VLMs are directly applied to the data without any finetuning or prompt optimization (the prompts are fixed beforehand).
    \item \textbf{Prompt-optimized evaluation:} The prompts are optimized by observing the model outputs on a small subset of videos from held-out subjects (denoted by PO in Table~\ref{tab:stroke_cohort}). The models are then evaluated (without any finetuning) on a  completely independent set of test subjects (denoted by TC in Table~\ref{tab:stroke_cohort}).
\end{itemize}

\paragraph{Data acquisition.}

The data consist of videos recorded from two different camera angles, at a resolution of $1088$ x $704$ pixels and $60$ or $100$ frames per second. The subjects were recorded in two different contexts: (1) rehabilitation based on activities of daily living and (2) impairment quantification via the FMA.

The rehabilitation videos include nine activities: brushing teeth, combing hair, applying deodorant, drinking water, washing face, eating, putting on/taking off glasses, and performing repetitive target-directed movements---moving a toilet paper roll horizontally on a table (radial tabletop task, RTT) and vertically on a transparent shelf with different shelf heights (shelf) (see App.~\ref{app:activity_videos} for a more detailed description). Each subject performed each activity for $3-5$ trials. 

In the FMA videos, subjects performed one of $33$ movement items under the supervision of a trained expert. Each item is scored from $0$ to $2$. The aggregate FMA score is a number between $0$ (maximally impaired) and $66$ (healthy). A more detailed description of the FMA is provided in App.~\ref{app:fma_videos}.

%%%%%%%%%%%%%%%%%%%%%%%%%%%%%%%%%%%%%%%%%%%%%%%%%%%%%%%

\paragraph{(A) VLMs accurately classify high-level activities.}

\begin{figure}[]
    \centering
\begin{tabular}{  
 >{\centering\arraybackslash}m{0.485\linewidth} >{\centering\arraybackslash}m{0.5\linewidth}  }    %\includegraphics[width=0.495\linewidth]
 Independent evaluation & Prompt-optimized evaluation \\
\includegraphics[width=\linewidth]{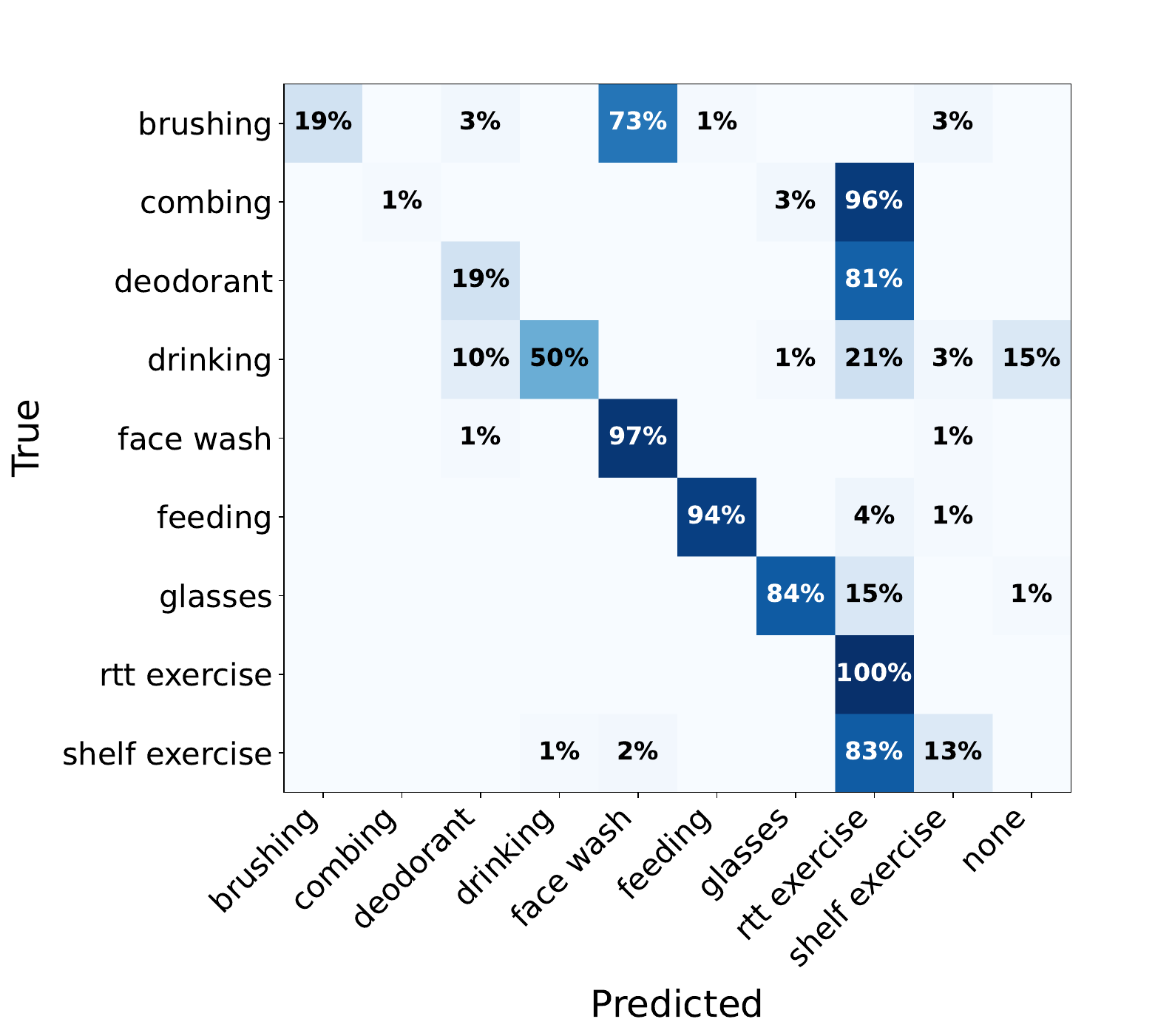}
    &
\includegraphics[width=\linewidth]{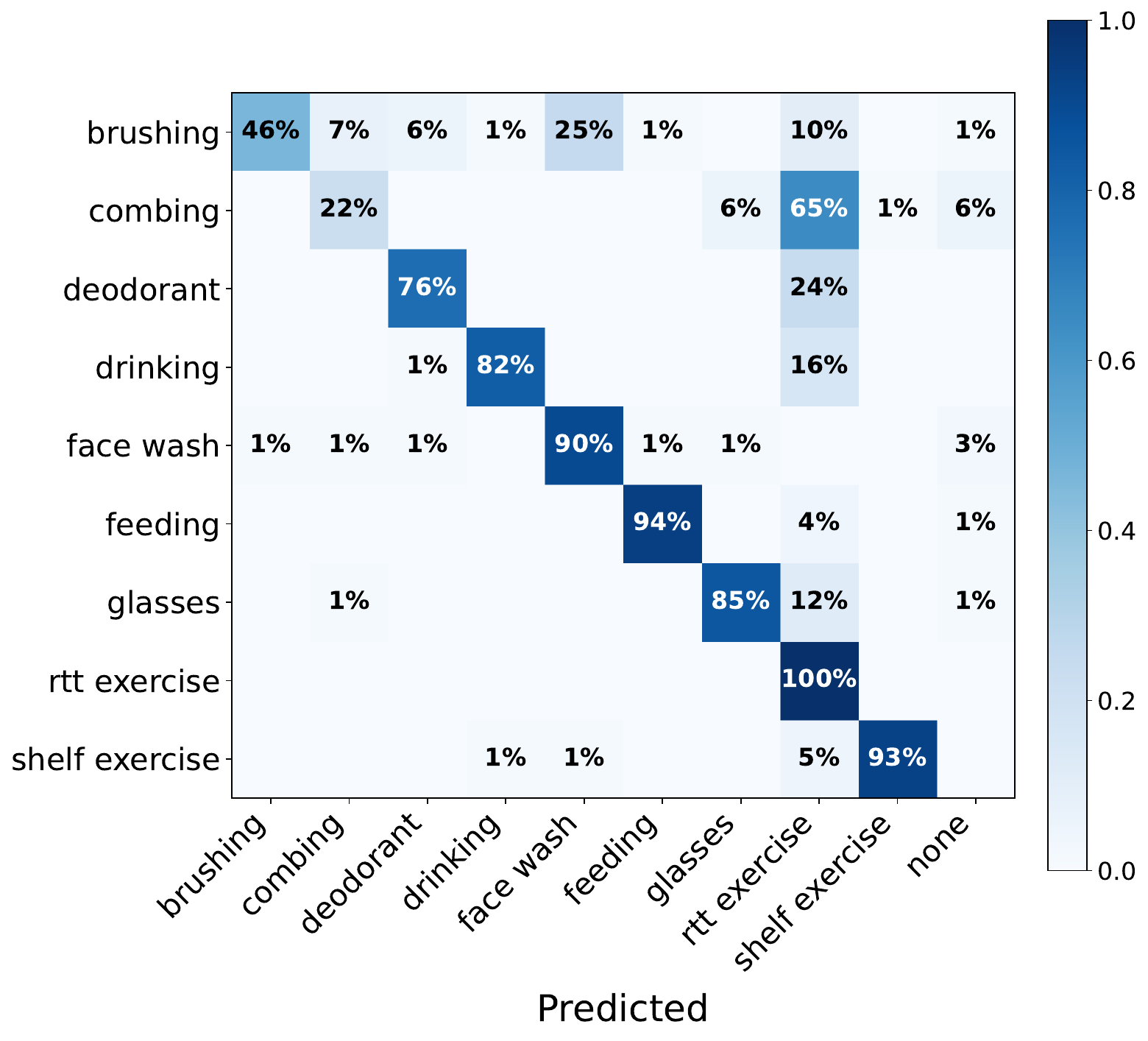}
    \end{tabular}
    %\hspace{0.3cm}
    \caption{\textbf{VLMs accurately classify high-level activities.} Shown are confusion matrices for activity identification with Qwen2.5-VL-7B-Instruct ($N=640$ videos). Each cell shows the fraction of samples from a true activity (row) that were predicted as a given activity (column). \textbf{Left:} Independent evaluation, where the VLM prompting is based on a pre-existing description of the activities. \textbf{Right:} Prompt-optimized evaluation, where the VLM prompts are tailored to Qwen2.5-VL-7B-Instruct using videos from $2$ held-out control subjects (see Methods). The optimized prompts achieve higher accuracy, as indicated by the dark diagonal cells. The most frequent mistakes are confusing face wash and RTT with the brushing and combing activities, respectively, possibly due to the small object sizes.}
    \label{fig:identification-cm}
\end{figure}

As a proof of concept of the ability of VLMs to understand the semantic content in rehabilitation videos, we consider the problem of identifying rehabilitation activities. As described in Figure~\ref{fig:fig1} and the Methods section, we perform activity recognition by providing eight uniformly sampled frames from each video, together with a prompt that describes the nine activities and asks the VLM to identify which one is depicted in the video. We applied this procedure to $640$ videos featuring $18$ healthy subjects and $51$ stroke patients (``(A) Activity Identification" on the right of Table~\ref{tab:stroke_cohort}) using the Qwen2.5-VL-7B-Instruct model.

The left confusion matrix in Fig.~\ref{fig:identification-cm} shows the results for independent evaluation, where the prompts to the VLM are based on a pre-existing description of the activities. The VLM achieves low accuracy ($53.4\%$ on average), but substantially better than predicting the majority class ($13.4\%$; slightly more than $1/9$ because some subjects perform the RTT and shelf activity twice, once per side). The VLM tends to over-predict RTT for the combing, deodorant, and shelf activities, possibly because they take place in the same workspace as RTT and often involve similar or small objects.

The right confusion matrix in Fig.~\ref{fig:identification-cm} shows the results for prompt-optimized evaluation, where the prompts to the VLMs are chosen using videos from two held-out control subjects, as described in the Methods section. Optimized prompting improves performance substantially, achieving an average accuracy of $77.5\%$. Mistakes primarily arise from  mis-classifying teeth brushing as face wash and combing as RTT, again likely due to the small sizes of the objects involved. Interestingly, the optimized prompt substantially boosts the performance of Qwen2.5-VL-7B-Instruct, but not that of the larger models in the Qwen family ($56.4\%$ for Qwen2.5-VL-32B-Instruct and $64.7\%$ for Qwen2.5-VL-72B-Instruct), illustrating that prompt optimization is specific to a given model. In conclusion, we find that VLMs are able to robustly classify activities with relatively high accuracy from a small number of frames, as long as the prompts are optimized using a limited number of held-out data.

\paragraph{(B) VLMs are able to detect subtle motions to some extent, but not enough for precise dose quantification.}

\begin{table}[]
  \centering
   \caption{\textbf{In dose quantification, VLMs perform only slightly better than a baseline that is independent from the visual input.} The table shows the edit score (ES), action error rate (AER), and relative counting error (RCE) of 15 VLMs with different sizes. Arrows indicate the direction of better performance. \colorbox{green!20}{Green} indicates the best result (including ties) among all models, and \colorbox{blue!20}{blue} the second-best result. Cells show mean $\pm$ $1$ sem. The best models are barely better than the Markov baseline, which only relies on transition statistics, and are very far from the Omniscient baseline, which leverages the ground-truth primitive sequences. The bottom of the table reports the performance of Qwen2.5-VL-32B-Instruct with engineered inputs. Evaluation was completely independent, without any prompt optimization.}
   \vspace{0.2cm}
\begin{tabular}{@{}lccc@{}}
\toprule
Model & Edit Score $\uparrow$ & Action Error Rate $\downarrow$ & Rel. Counting Error $\downarrow$ \\
\midrule
\textbf{Baselines} & & & \\
Markov (does not rely on videos) & 46.32 ± 0.91 & 0.71 ± 0.08 & 0.63 ± 0.08 \\
Omniscient & 88.47 ± 0.87 & 0.12 ± 0.01 & 0.11 ± 0.01 \\
\midrule
%%%%%%%%%%%%%%%%%%%%%%%%%%%%%%%%%%%%%%%%%%%%%%%%%%%%%%%%%%%%%%%%%%%%%%%%%%%%%
InternVL3-78B & \cellcolor{green!20}\textbf{49.83 ± 1.21} & 0.68 ± 0.09 & 0.73 ± 0.09 \\
InternVL3.5-2B & 34.20 ± 1.98 & 0.81 ± 0.12 & 0.80 ± 0.12 \\
InternVL3.5-8B & 8.17 ± 1.24 & 0.92 ± 0.01 & 0.93 ± 0.01 \\
InternVL3.5-38B & 40.20 ± 1.49 & 0.68 ± 0.03 & 0.80 ± 0.03 \\
InternVL3.5-30B-A3B & 25.54 ± 1.94 & 0.78 ± 0.02 & 0.87 ± 0.03 \\
LLaVA-NeXT-Video-7B & 46.18 ± 1.76 & \cellcolor{blue!20}\textbf{0.65 ± 0.07} & \cellcolor{blue!20}\textbf{0.68 ± 0.07} \\
LLaVA-NeXT-Video-72B & 23.08 ± 1.65 & 0.78 ± 0.02 & 0.97 ± 0.03 \\
LLaVA-OneVision-0.5B & 20.97 ± 1.44 & 0.80 ± 0.01 & 0.96 ± 0.02 \\
LLaVA-OneVision-7B & 42.07 ± 1.31 & 0.69 ± 0.04 & 0.78 ± 0.05 \\
LLaVA-OneVision-72B & 38.52 ± 1.93 & 0.72 ± 0.07 & 0.76 ± 0.08 \\
NVILA-8B & 36.98 ± 2.09 & 0.75 ± 0.09 & 0.76 ± 0.09 \\
NVILA-15B & 24.92 ± 1.21 & 0.79 ± 0.04 & 0.95 ± 0.04 \\
Qwen2.5-VL-7B-Instruct & 34.75 ± 1.73 & 0.69 ± 0.02 & 0.77 ± 0.02 \\
Qwen2.5-VL-32B-Instruct & \cellcolor{blue!20}\textbf{46.69 ± 1.62} & \cellcolor{blue!20}\textbf{0.65 ± 0.06} & \cellcolor{green!20}\textbf{0.65 ± 0.06} \\
Qwen2.5-VL-72B-Instruct & 44.88 ± 1.58 & \cellcolor{green!20}\textbf{0.58 ± 0.02} & \cellcolor{green!20}\textbf{0.65 ± 0.02} \\
%%%%%%%%%%%%%%%%%%%%%%%%%%%%%%%%%%%%%%%%%%%%%%%%%%%%%%%%%%%%%%%%%%%%%%%%%%%%%
\midrule
\textbf{Engineered inputs} \\(Qwen2.5-VL-32B-Instruct) & & & \\
Cropping & 48.28 ± 1.53 & 0.61 ± 0.04 & 0.62 ± 0.04 \\
Contextual Prompting & 48.32 ± 1.61 & 0.73 ± 0.08 & 0.83 ± 0.08 \\
Cropping \& Contextual Prompting & 53.18 ± 1.50 & 0.70 ± 0.05 & 0.76 ± 0.05 \\

\bottomrule
\end{tabular}
\label{tab:smc_llms}
\end{table}

\begin{figure}[t!]
\centering
\begin{minipage}{0.48\linewidth}
  \centering
  \subcaption*{(a)}
  \includegraphics[width=\linewidth]{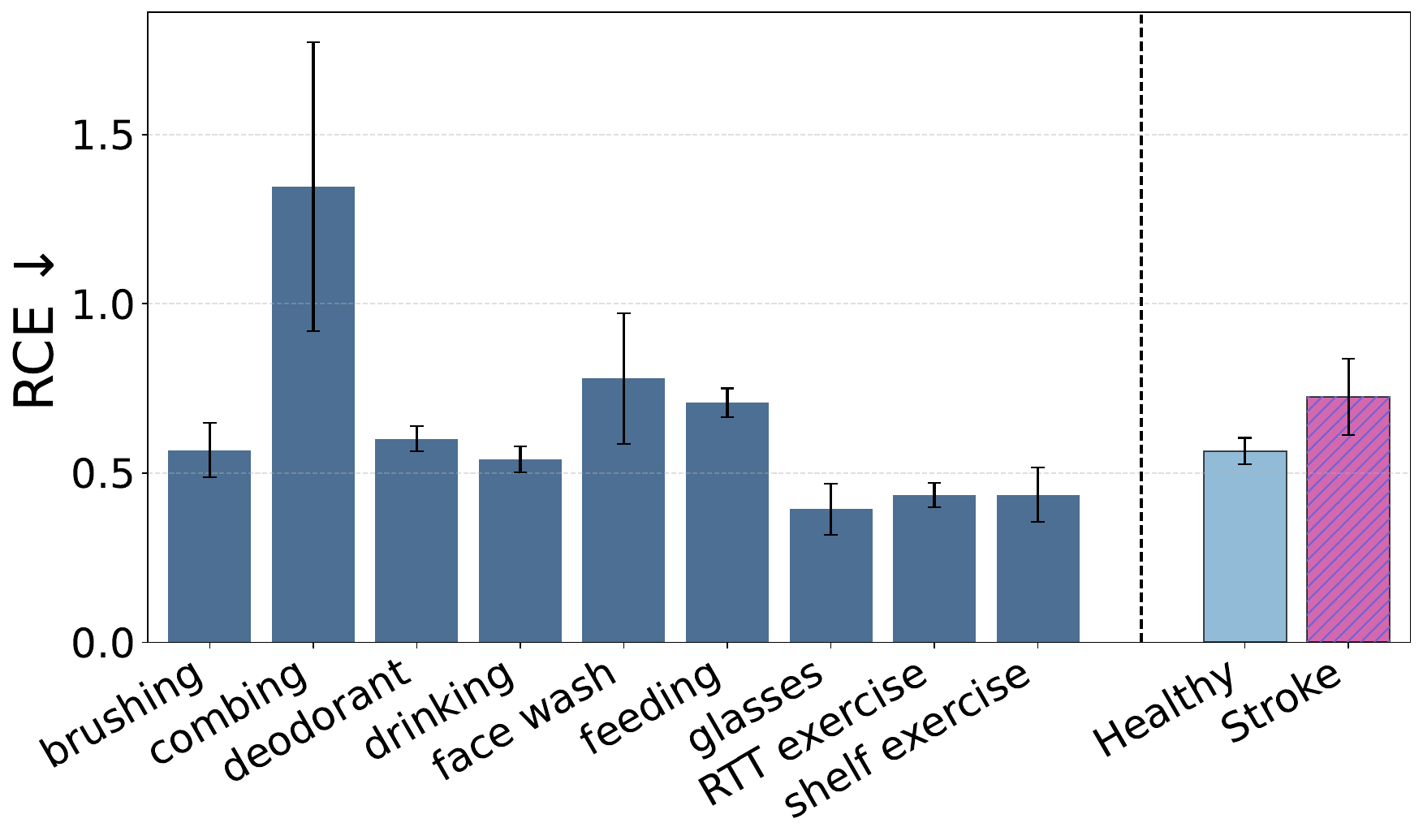}
\end{minipage}\hfill
\begin{minipage}{0.48\linewidth}
  \centering
  \subcaption*{(b)}
  \includegraphics[width=\linewidth]{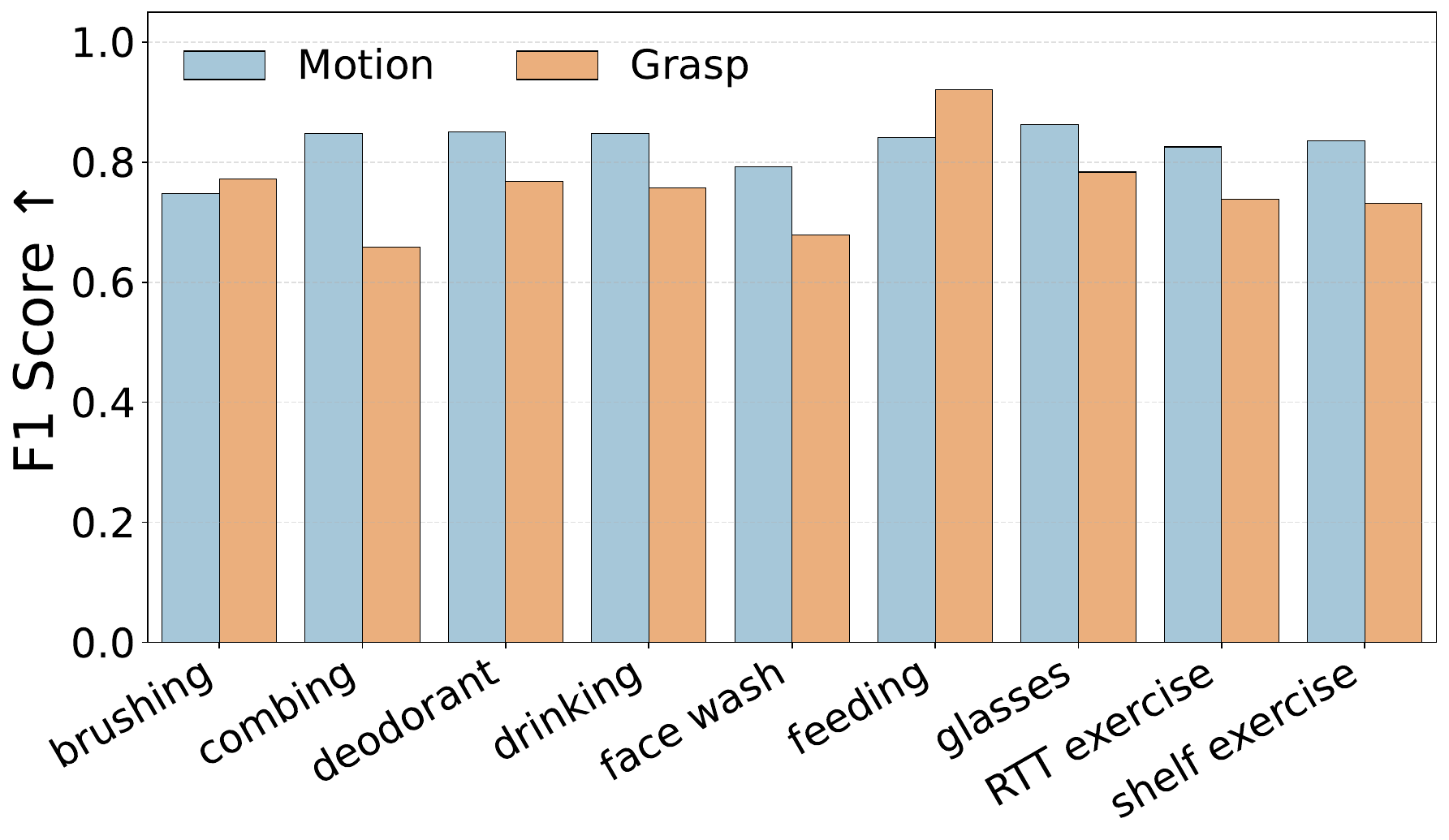}
\end{minipage}

\vspace{0.3em}

\begin{minipage}{0.48\linewidth}
  \centering
  \subcaption*{(c)}
  \includegraphics[width=\linewidth]{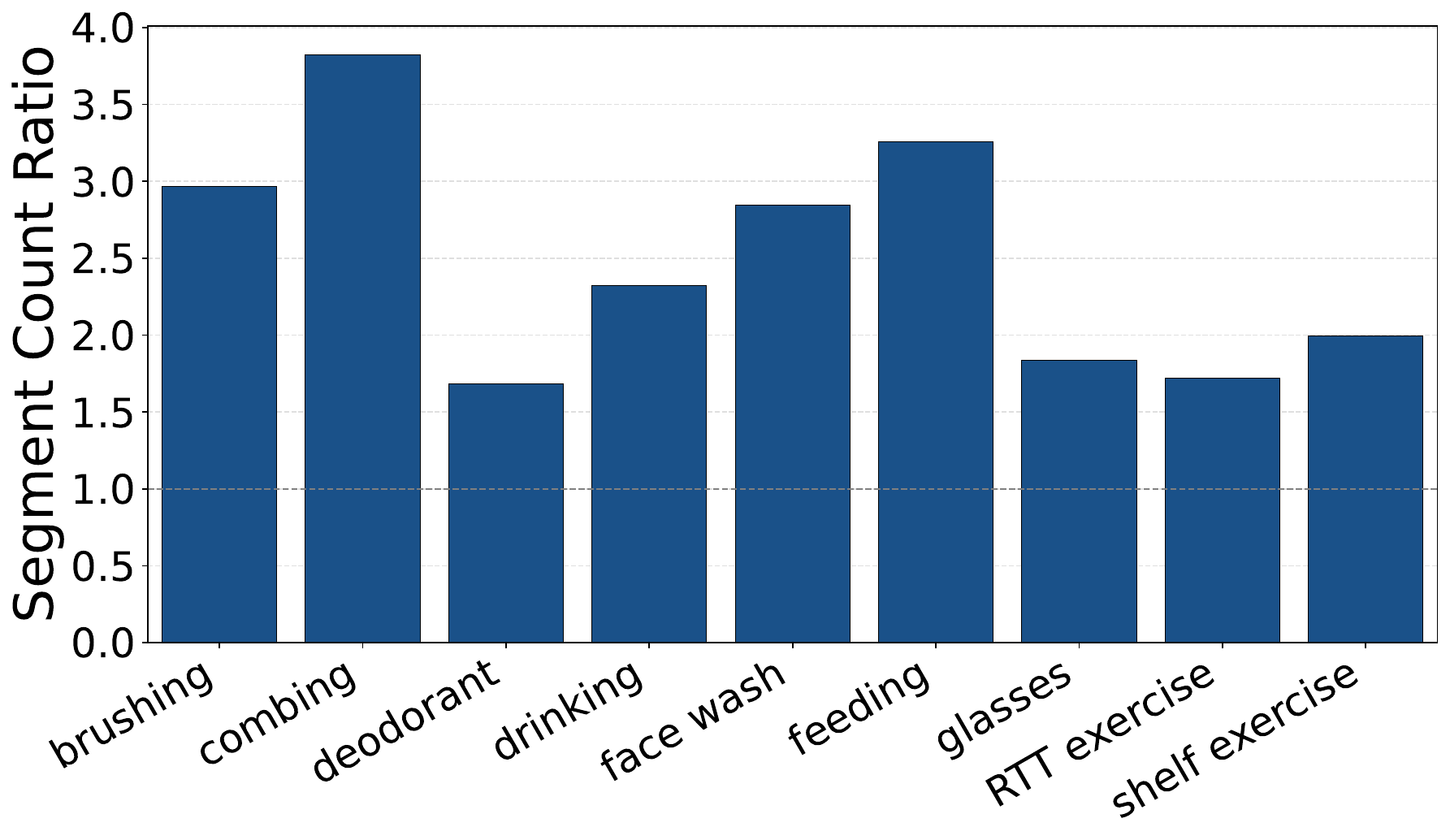}
\end{minipage}\hfill
\begin{minipage}{0.48\linewidth}
  \centering
  \subcaption*{(d)}
  \includegraphics[width=\linewidth]{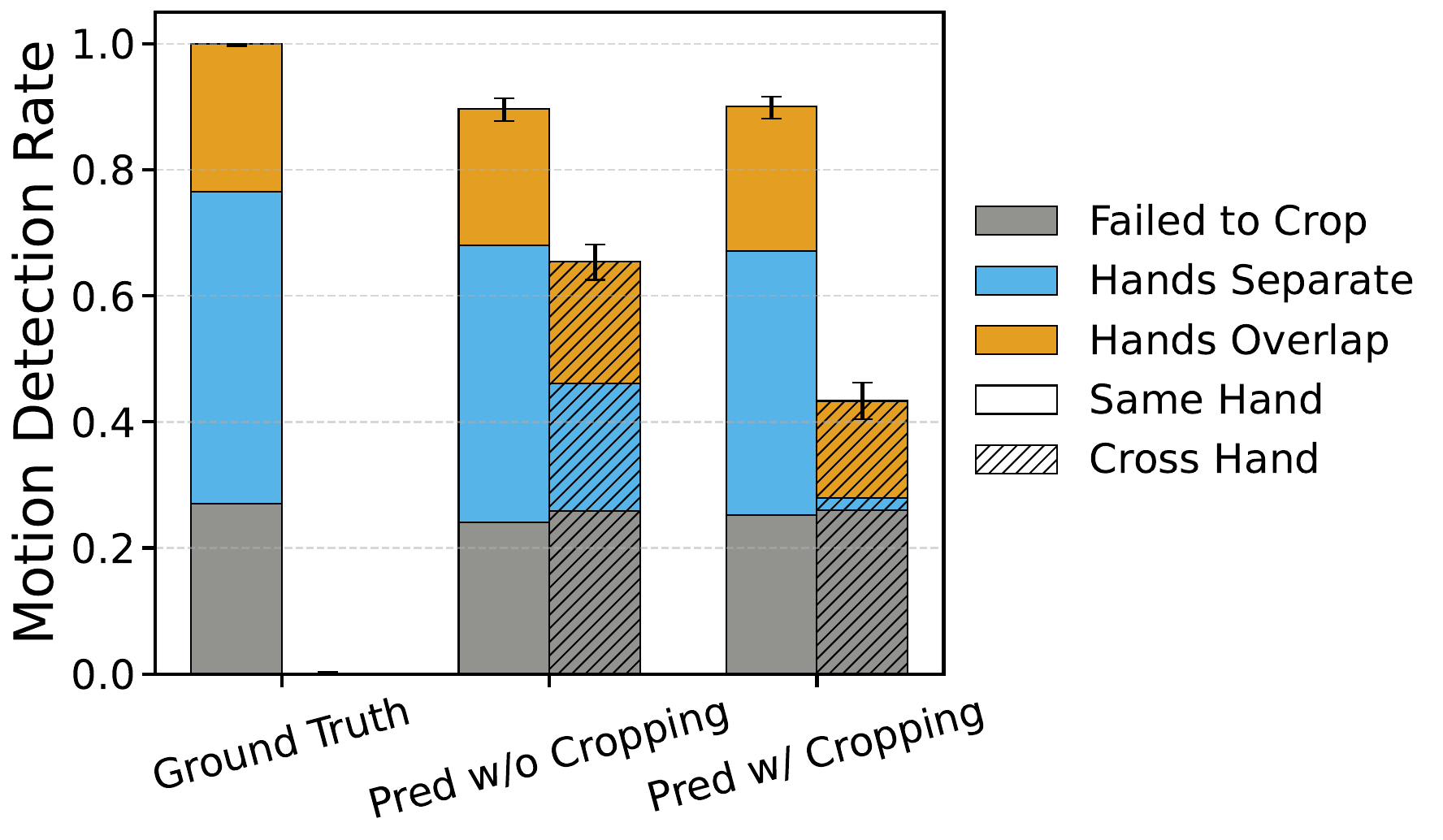}
\end{minipage}

\caption{\textbf{VLMs detect motion and grasp with relatively high accuracy (albeit insufficient for precise dose quantification), but confuse left and right. } The graphs provide a  nuanced analysis of one of the best-performing models: Qwen2.5-VL-32B-Instruct with cropping. Evaluation was completely independent, with no prompt optimization. \textbf{(a)} The relative counting error (RCE $\downarrow$) is high ($\approx 50\%$ for most activities) and is worst for \textit{combing}. Healthy subjects show slightly lower RCEs compared to stroke patients. The bars represent mean $\pm$ $1$ sem. \textbf{(b)} The F1 score of the predicted motion and grasp is high across all activities, indicating accurate detection. \textbf{(c)} The model switches predictions between consecutive segments too frequently---especially for combing.
\textbf{(d)} We isolated segments where only one hand was active. When instructed to track a particular hand, a VLM should detect motion $100\%$ of the time for the moving hand and $0\%$ for the still hand. Empirically, the VLM mistakenly classifies the still hand as moving in $>60\%$ of cases, presumably due to the motion in the other hand.
Cropping partially alleviates this issue, but can fail and does not help when the hands overlap. The error bars show $95\%$ binomial proportion confidence intervals.}
\label{fig:vlm_analysis}
\end{figure}

We formulate the task of measuring rehabilitation dosage as identification of fine-grained motions called \textit{functional primitives}. These primitives are basic building blocks of upper extremity (UE) motion \citep{schambra2019taxonomy}, and can be aggregated to produce counts that quantify rehabilitation dose. The functional primitives are \textit{reach} (move to grasp or touch a target object), \textit{reposition} (move proximate to a target object, e.g., the initial neutral spot), \textit{transport} (move a grasped target object), \textit{stabilize} (hold a target object still), and \textit{idle} (stand at the ready near a target object). % See Table~\ref{tab:prim_def} in Methods for additional details.

To process a video, we divide it into $0.533$-second segments, extract eight frames uniformly per segment, and use the VLM to identify a \textit{single} primitive for each segment. Afterwards, we concatenate the predictions and \textit{de-duplicate} consecutive identical primitives. A limitation of this approach is that some primitives are shorter than the duration of the segment ($0.533$ s), so in principle this could lead to under-counting, but these short-duration primitives are rare (see Fig.~\ref{fig:activity_metadata_plot}).

We design the VLM prompt by exploiting the fact that the primitives can be determined based on the presence of significant motion and grasp (see Table~\ref{tab:prim_def} in Methods). For each segment, the model is prompted twice to obtain binary predictions for motion and grasp. These results are then post-processed into a final primitives sequence, as explained in the Methods section. We term this approach \emph{Decomposed Prompting}. Further ablations concerning prompting, the segment duration, and the number of frames per segment are provided in App.~\ref{app:further_results_prims_id}.

For evaluation, we have access to per-frame ground-truth primitive labels. These annotations pertain to a target side (left or right) and were meticulously labeled by trained annotators with Cohen's kappa $\geq0.96$ between the labelers and the expert. We obtain the predicted sequence for the target side and compare the predicted and ground-truth sequences  using three metrics. The \textit{edit score} (ES, range $0$ to $100$, higher is better) and \textit{action error rate} (AER, range $\geq 0$, lower is better) quantify the similarity between two sequences. 
The \textit{relative counting error} (RCE, range $\geq 0$, lower is better) measures the error in rehabilitation dose quantification. It equals the sum of the counting errors of the five primitives normalized by the ground truth sequence length. See Methods for a formal definition of these metrics.

We evaluated the proposed dose-quantification method on a test set of $90$ videos comprising nine activities for each of five control subjects and five mild/moderate stroke patients (see ``(B) Dose Quantification" on the right of Table~\ref{tab:stroke_cohort}). The total number of primitives in the test dataset was $4,059$. The evaluation was fully independent, without any prompt optimization. Two baselines provide context for VLM performance. (1) \textit{Markov} is based only on the first-order transition statistics for motion and grasp, and provides a strong baseline that does not have access to the videos.
(2) \textit{Omniscient} has access to the ground-truth annotations of motion and grasp for each segment. It serves as an upper bound on achievable performance under our assumptions regarding the dependence of each primitive on motion and grasp, and the constraint that only one primitive is present per segment. See the Methods section for additional details on these baselines.

Table~\ref{tab:smc_llms} shows the results of applying 15 different VLMs for dose quantification using the proposed framework. The relative counting error is 65\% or greater for all models. In fact, the best models achieve only slightly better performance than the Markov baseline, which does not take into account the video content. Using one of the best models, Qwen2.5-VL-32B-Instruct, we evaluated the effect of visual curation (cropping the input frames around the relevant upper extremity) and contextual prompting (see Methods for more details). Cropping boosts performance slightly, while contextual prompting does not.

Figure~\ref{fig:vlm_analysis} provides a more detailed analysis of one of the best-performing models: Qwen2.5-VL-32B-Instruct with cropping. Figure~\ref{fig:vlm_analysis}(a) shows that the \textit{relative counting error} (RCE) is close to $50\%$ across activities, with extremely poor performance for combing. The aggregated RCE is lower for healthy subjects than for stroke patients, but still above $50\%$. Figure~\ref{fig:vlm_analysis}(b) reports the performance of the VLM for motion and grasp detection \textit{at the segment level.}
The F1 score between the predictions and the ground-truth motion/grasp labels is  high for most activities (above 0.7 for grasp and above 0.8 for motion), indicating that the VLM outputs are mostly correct. However, they are not accurate enough to result in precise primitive counts. Figure~\ref{fig:vlm_analysis}(c) shows that the VLM primitive estimates are quite unstable, constantly switching between primitives, which results in over-segmentation: $1.5$ to $4$ times more segments are predicted compared to the number of ground-truth segments after de-duplication.

%Separately, 
An interesting limitation of VLMs is that they fail to follow instructions that require differentiating between the left and right hand, despite careful prompt design (see App.~\ref{app:prim_id_prompts}). In Figure~\ref{fig:vlm_analysis}(d), we report the results of an experiment based on segments where one hand is actively moving and the other is not. We prompted the VLM once for each hand, asking whether that hand was currently moving. Ideally, a model should detect motion $100\%$ of the time for the active hand and $0\%$ of the time for the non-active hand. However, we found that the model declared the non-active hand to be \textit{moving more than $60\%$ of the time}. Cropping partially mitigates this issue, but fails for some videos and remains ineffective when the hands overlap. More details on this experiment are provided in the Methods section.

\paragraph{(C) With  prompt optimization and post-processing, VLMs can quantify rehabilitation dosage for structured activities to some extent.}

\begin{table}
\captionsetup{width=\textwidth}
\caption{\textbf{Prompt optimization, cropping and post-processing improve dose quantification.} Results of \emph{PRIM-RS} (Pose-Refined promptIng Module---RTT/shelf), which incorporates optimized prompting, cropping and post-processing, compared to \emph{Decomposed Prompting} (based on non-optimized predefined prompts) for a curated dataset of $38$ videos comprising the RTT and shelf tasks from five subjects in each cohort---healthy, mild, moderate, and severe (two severe patients did not complete the shelf task). We use the aerial camera angle for the shelf task and the side view closer to the hand of interest for the RTT task. PRIM-RS improves upon Decomposed Prompting for the three metrics. Cropping around the hand of interest and careful post-processing provide gains in performance. Arrows indicate the direction of better performance. Cells show mean $\pm$ $1$ sem.}
\vspace{0.3cm} 
\begin{tabular}{lcccr}
\toprule
Prompt Method & Edit Score $\uparrow$ & Action Error Rate $\downarrow$ & Rel. Counting Error $\downarrow$ \\
\midrule
PRIM-RS & 67.75 ± 3.26 & 0.42 ± 0.05 & 0.40 ± 0.06 \\
-- w/o cropping & 61.37 ± 3.26 & 0.46 ± 0.05 & 0.45 ± 0.05 \\
-- w/o post-processing & 50.61 ± 2.90 & 1.04 ± 0.26 & 1.01 ± 0.26 \\
-- w/o cropping/post-processing & 44.94 ± 2.81 & 0.79 ± 0.08 & 0.87 ± 0.09 \\
Decomposed Prompting & 47.29 ± 2.61 & 1.07 ± 0.28 & 1.20 ± 0.31 \\
\bottomrule
\end{tabular}
\label{tab:prim_v_dp}
\end{table}

\begin{figure}
    \centering
    \includegraphics[width=\linewidth]{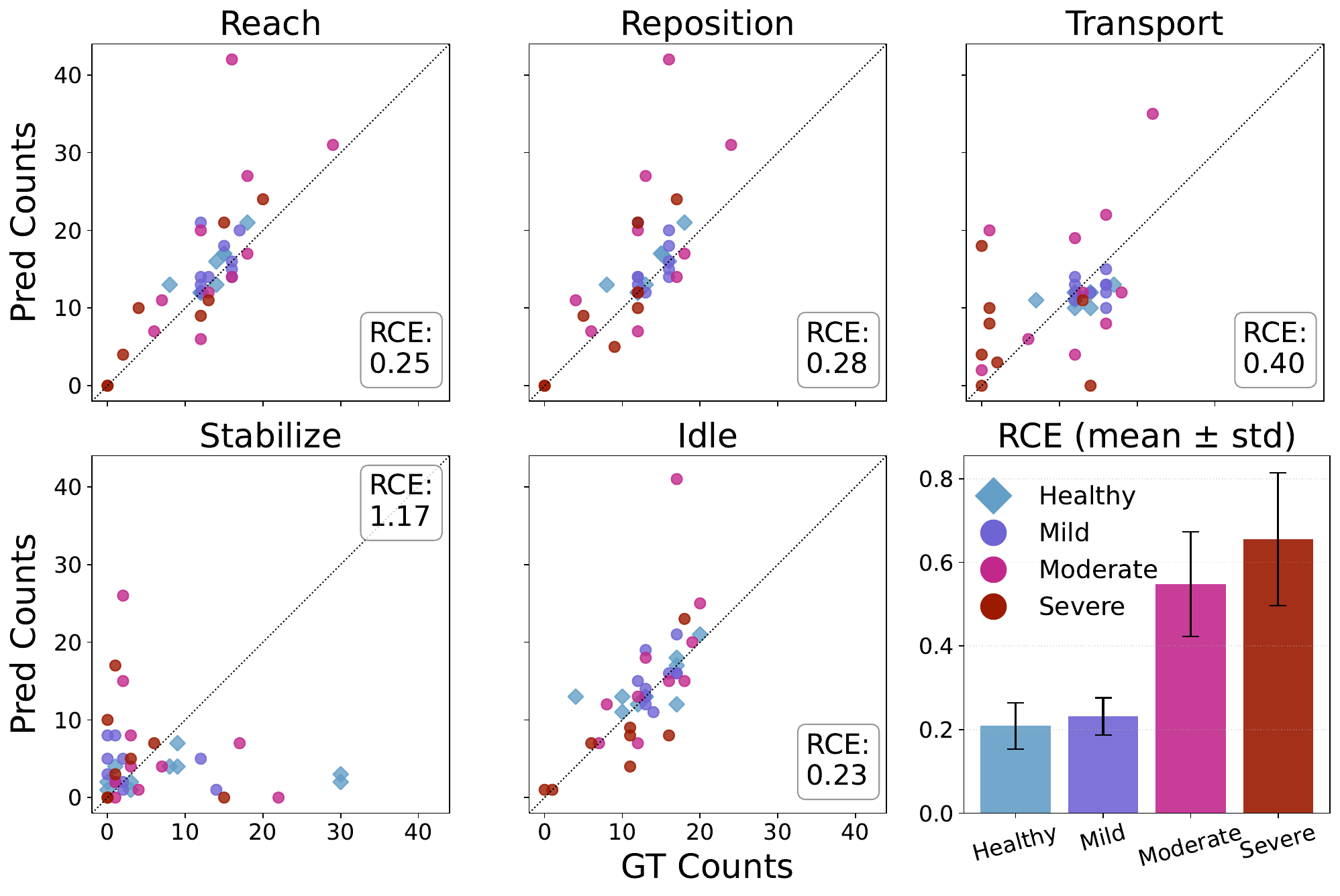}
    \caption{
    \textbf{
    With prompt optimization and post-processing, VLMs can quantify rehabilitation dosage for structured activities to some extent.} The first five panels shows scatterplots comparing the predicted and ground-truth counts for $38$ rehabilitation videos corresponding to two structured tasks (radial table top and shelf). Predictions are obtained using the proposed PRIM-RS method, which provides optimized prompts to the VLM Qwen2.5-VL-32B-Instruct and applies post-processing to its output. The relative counting error (RCE) is primitive-specific, measuring the counting error for each primitive normalized by its total number of ground-truth instances in the $38$-video dataset. The RCE is moderately low at $\approx25\%$ for \textit{reach}, \textit{reposition}, and \textit{idle}. It is subpar for \textit{transport} and \textit{stabilize}, illustrating the difficulty of differentiating these primitives. The bottom-right plot shows the average RCE across subjects with different impairment levels: the performance degrades as the severity level increases.}
    \label{fig:rtt_shelf_counts}
\end{figure}

The radial table top (RTT) and shelf tasks are structured activities involving repetitive motions that are relatively similar among subjects. We propose a pipeline for dose quantification from videos of these activities, which combines a VLM (Qwen2.5-VL-32B-Instruct) with a pose model. The pipeline is optimized using held-out videos of two control subjects and one severe patient. We compare this pipeline, named Pose-Refined promptIng Module---RTT/Shelf (\emph{PRIM-RS}), with the \emph{Decomposed Prompting} strategy of the previous subsection using a test set of videos featuring five subjects from each cohort (see ``(C) Dose Quant. RTT/Shelf" on the right of Table~\ref{tab:stroke_cohort}). 

The results in Table~\ref{tab:prim_v_dp} demonstrate that activity-specific optimizations boost performance. Most notably, the post-processing step in PRIM-RS, which involves smoothing to prevent over-segmentation, is critical. Cropping around the desired hand yields additional gains.

Figure~\ref{fig:rtt_shelf_counts} displays the dose quantification results for PRIM-RS. For the \textit{reach}, \textit{reposition}, and \textit{idle} primitives, the \textit{per-primitive counting error} (see caption) is moderately accurate at approximately $25\%$. This error is partly driven by a single outlier video in which PRIM-RS fails substantially. In contrast, performance for \textit{transport} and especially \textit{stabilize} is subpar, reflecting the difficulty of distinguishing between these primitives. This mainly stems from differences in movement speed across severity levels, and from the fact that \textit{stabilize} durations can be extremely brief. The bottom right of Figure~\ref{fig:rtt_shelf_counts} shows the mean performance for subjects with different impairment levels. PRIM-RS performs considerably better (close to 20\% RCE) for healthy subjects and mild patients, but its performance deteriorates for moderate and severe patients, whose movements are irregular and frequently idiosyncratic.

\paragraph{(D) VLMs are not able to perform impairment quantification.}

\begin{figure}[t]
    \centering
    \includegraphics[width=0.48\linewidth]{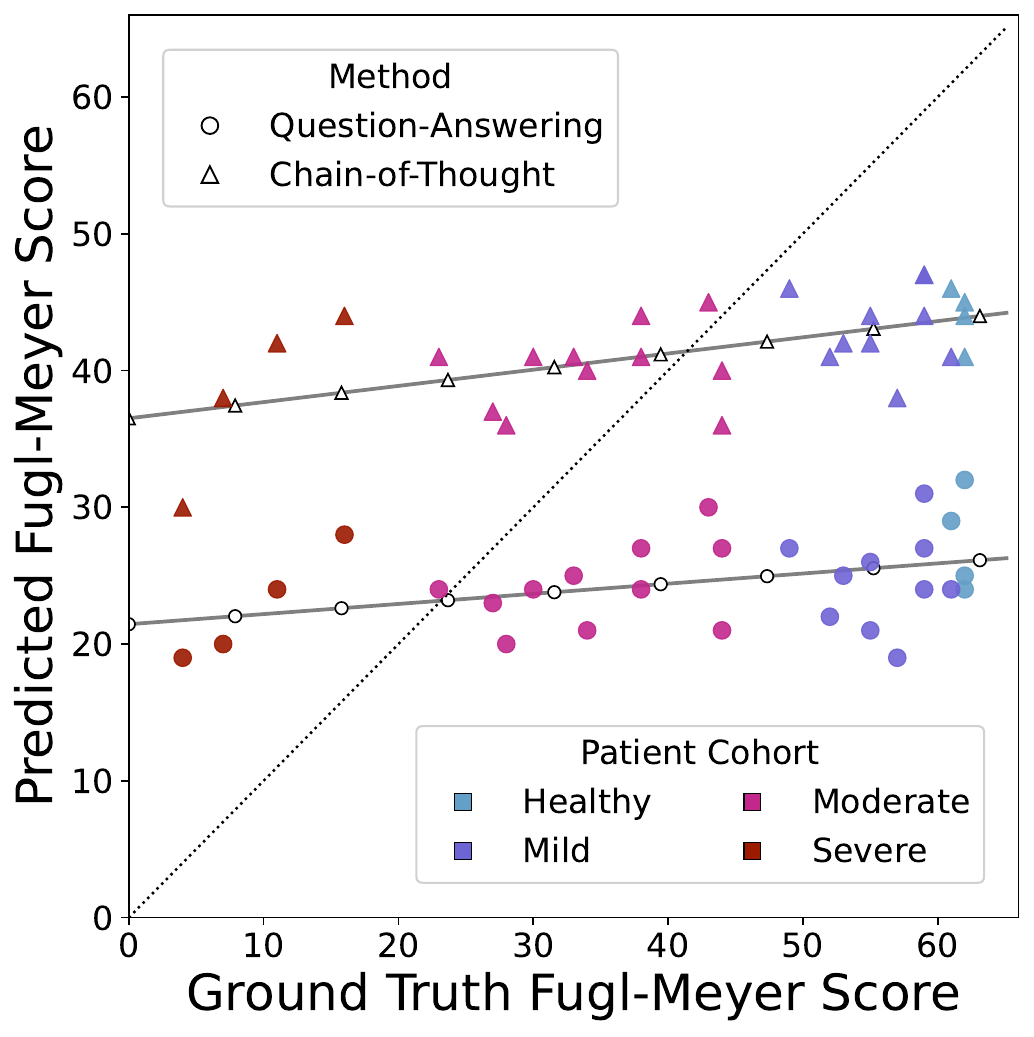}
    \hfill
    \includegraphics[width=0.48\linewidth]{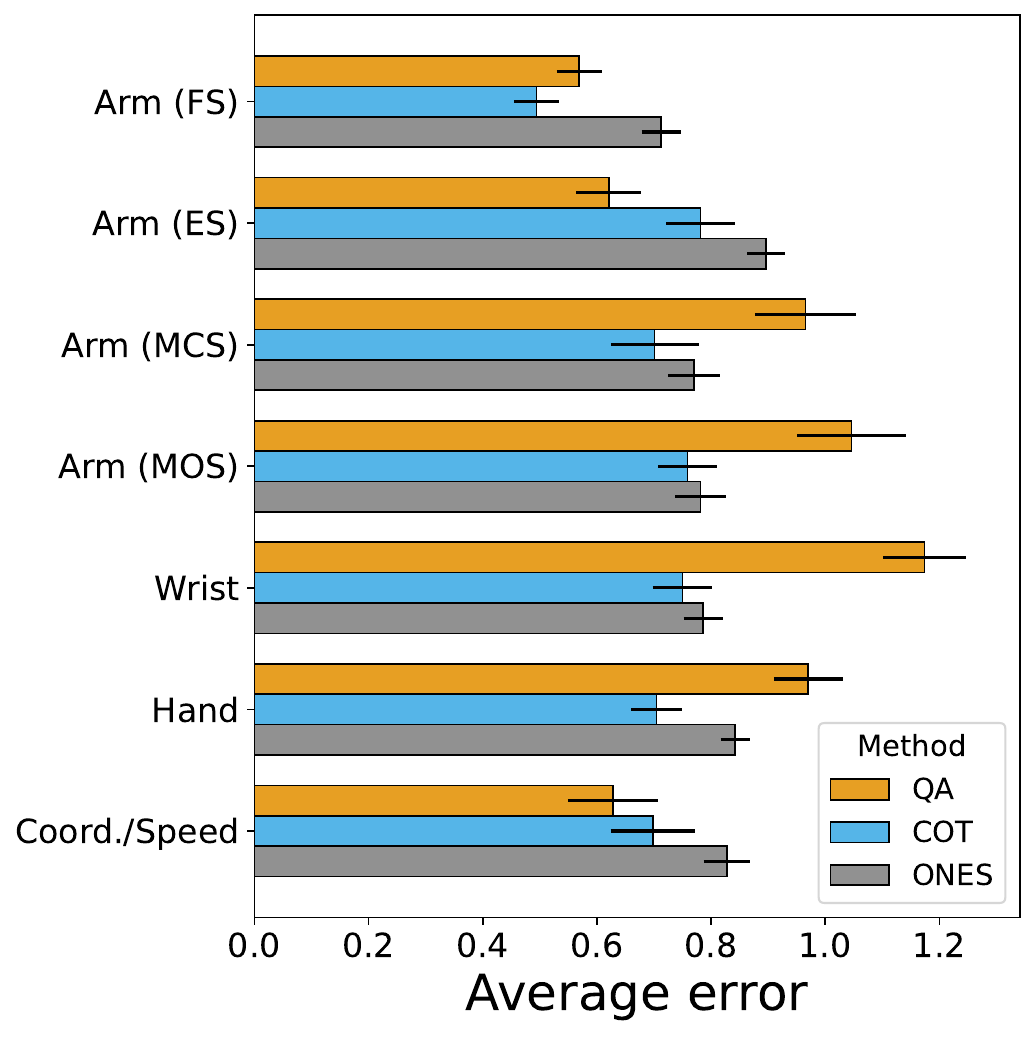}
    \caption{\textbf{VLMs fail at impairment quantification.} \textbf{Left: } The scatterplot shows the ground-truth Fugl-Meyer score assessed by a trained human expert against the predicted Fugl-Meyer score by the VLM Qwen2.5-VL-72B-Instruct. The points would lie along the dotted diagonal line for a model that matches human rating. Instead, for two different prompting methods, the predicted Fugl-Meyer score is nearly constant across severity levels. \textbf{Right: } The bar plots show the average error of the VLM for different subsections of the Fugl-Meyer assessment, again using two prompting methods. For comparison, ONES displays the average error for a model that only predicts $1$. The subsections are, from top to bottom, Arm---Flexor Synergy, Arm---Extensor Synergy, Arm--Movement Combining Synergy, Arm---Movement Out of Synergy, Wrist, Hand, and Coordination/Speed. Both methods perform comparatively to the non-informative ONES baseline. Bars show mean $\pm$ $1$ sem. %\textit{Note: we do not evaluate the \textbf{Reflexes} section. This section has two items, hence why the maximum ground-truth Fugl-Meyer score is $62$.}
    }
    \label{fig:impairment_assessment}
\end{figure}

As illustrated in Figure~\ref{fig:fig1}, we leverage the Fugl-Meyer assessment (FMA) to use VLMs for impairment quantification. The assessment consists of $33$ items or questions  about a subject's movements. We feed the questions as prompts to the VLM, requesting it to classify the movement quality as $0$ (worst), $1$, or $2$ (best).

We evaluated the proposed framework on $899$ videos from a cohort of $28$ subjects (see ``(D) Impairment Quantification" on the right of Table~\ref{tab:stroke_cohort}). Evaluation was completely independent, with no prompt optimization. Two alternative prompting methods were applied. (1) \emph{Question-Answering (QA)} asks the VLM multiple binary questions regarding movement quality. (2) \emph{Chain-of-Thought (CoT)} \citep{wei2023chainofthoughtpromptingelicitsreasoning} provides all relevant context to the VLM in one prompt and asks it to give a final answer after making relevant observations. More details on these two prompting techniques can be found in the Methods section.

Figure~\ref{fig:impairment_assessment} shows the results for Qwen2.5-VL-72B-Instruct. The predicted Fugl-Meyer score in the left plot is essentially constant throughout the spectrum of severity levels, indicating that the VLM fails to quantify impairment. The right plot breaks down the average error for all patients and scoring items for different subsections of the FMA. Errors for the two methods are similar to a non-informative model that ignores the visual input and returns a score of $1$.

%%%%%%%%%%%%%%%%%%%%%%%%%%%%%%%%%%%%%%%%%%%%%%%%%%%%%%%%%%%%%%

%% file: sections/03_discussion.tex
\section*{Discussion}
\label{sec:discussion}

\begin{figure}
    \centering
    \includegraphics[width=\linewidth]{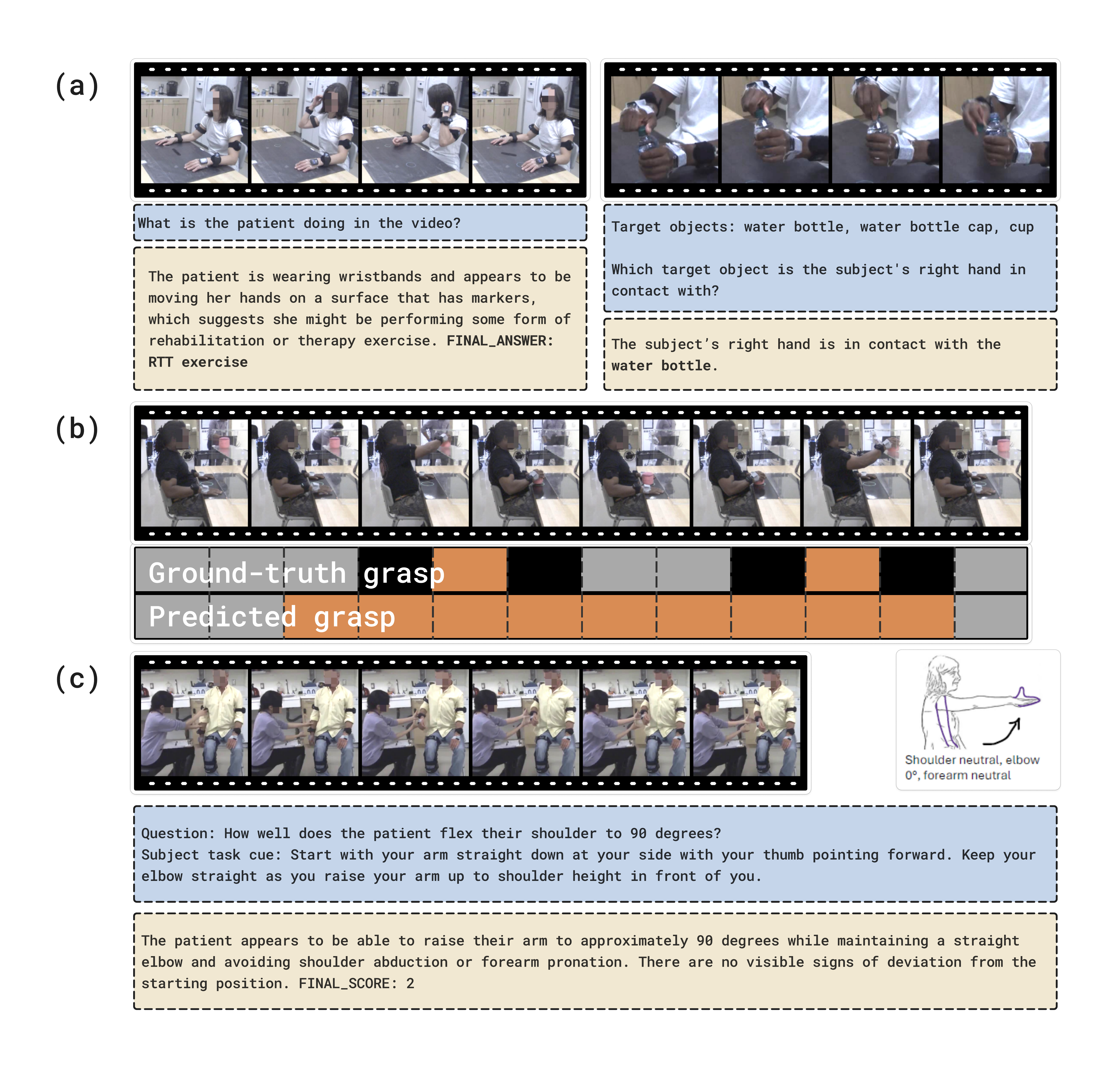}
    \vspace{-0.4em}
    \caption{\textbf{Failure modes of VLMs applied to stroke rehabilitation.} \textbf{(a) Large object bias:} Left: The model misclassifies a combing activity as an RTT exercise, likely driven by contextual cues from larger objects such as the black mat and wristbands. Right: The model exhibits hand attribution errors, potentially due to the left hand's interaction with a dominant object (water bottle). \textbf{(b) Overreliance on 2D semantics:} Timeline of a $6$\,s video with dotted lines marking $0.533$\,s segments. Colors indicate grasp state: gray (no grasp), orange (grasp), and black (mixed). The ground truth contains two distinct grasps. When queried, ``Is the subject's right hand grasping the pink object? Answer `Yes' or `No' directly." for each segment, the VLM misinterprets visual proximity as physical contact and fails to distinguish the two separate grasp events. \textbf{(c) Hallucination:} A patient with severe impairment attempts a shoulder flexion task (ground truth Fugl-Meyer score of $0$). For reference, the diagram on the right depicts successful completion of the task. The model hallucinates movement and incorrectly reports task success.}
    \label{fig:failure_modes}
\end{figure}

This study reaches a nuanced conclusion regarding the current capabilities of vision-language models (VLMs) for human motion understanding. 

On the positive side, after prompt optimization on a limited number of held-out subjects, VLMs identify high-level activities accurately, and can be engineered to perform fine-grained motion identification in highly structured contexts. Moreover, VLMs can detect the presence of substantial motion or grasp with moderate accuracy, even without prompt optimization. Remarkably, these capabilities emerge without any task-specific training or finetuning on similar data distinguishing VLMs from traditional machine-learning approaches that rely heavily on domain-specific supervision. This is particularly attractive for digital medicine applications, such as stroke rehabilitation, where annotated data is scarce.

On the negative side, VLMs are unable to capture the subtle kinematic details required for precise quantification of rehabilitation dose or motor impairment, even after careful engineering. This is supported by our quantitative results, and also by qualitative observations that reveal several critical failure modes (see Fig.~\ref{fig:failure_modes}):

\begin{itemize}
\item \emph{Object bias:} VLMs tend to focus on large objects; for instance, several combing videos were misclassified as RTT seemingly because the presence of a black mat led the model to infer ``some form of exercise or therapy."
\item \emph{Limited sensitivity to fine movements:} The models fail to detect wrist motions like twisting and rotating, leading to poor performance on activities such as face washing or feeding.
\item \emph{Overreliance on 2D semantics:} VLMs may infer a grasp when a hand merely hovers near an object, lacking the depth reasoning needed to recognize actual contact.
\item \emph{Hallucinations:} In some cases, the models fabricate motion, reporting that a severely impaired patient performs a task correctly when barely any movement occurs.
\end{itemize}

In addition, we identify a key methodological limitation: the relatively short context length of VLMs, which restricts the number of video frames that can be processed simultaneously. This results in oversegmentation. Motion blur, occlusion, and camouflage (when the object boundary melds with the immediate background) are common in hand–object interactions, so objects may intermittently appear or disappear from view, causing the VLM to rapidly alternate its predictions for grasp detection. Attempts at contextual prompting (i.e. embedding prior information in the textual input to the VLMs) proved ineffective. Limited temporal context also leads to inaccurate motion detection. For example, without temporal context about the movement speed of a patient, the VLM struggles to distinguish between the \textit{transport} and \textit{stabilize} primitives, as these depend strongly on the individual's pace.

%Human action understanding is a core problem in video understanding, since \textit{humans} are often the subject of interest in a video. As computer vision models have become more powerful in the past decade, benchmarks in human action understanding have progressively become more granular. Early action-classification datasets \citep{soomro2012ucf101dataset101human,kay2017kineticshumanactionvideo} have gradually evolved into benchmarks that demand models to reason over finer spatio-temporal structure and human motion \citep{gu2018avavideodatasetspatiotemporally,goyal2017something,shao2020finegym}. And now with the advent of VLMs that can link vision to language, the most recent benchmarks require even more nuanced, language-grounded understanding---including reasoning with multiple actors \citep{salehi2024actionatlasvideoqabenchmarkdomainspecialized}, understanding fine-grained kinematics within a video \citep{shangguan2024tomatoassessingvisualtemporal}, identifying subtle differences between videos of the same action \citep{burgess2025video}, and understanding the world from the egocentric point of view \citep{grauman2022ego4dworld3000hours}.

Our findings suggest that achieving the level of human motion understanding required for data-driven stroke rehabilitation and other video-based digital health applications will necessitate both new datasets and methodological advances.
On the data side, curating high-quality, multi-site datasets with precise annotations of subtle human movements is essential. 
Encouraging progress has been made on this front, with recent benchmarks revealing qualitative limitations of video analysis with VLMs \citep{shao2020finegym,salehi2024actionatlasvideoqabenchmarkdomainspecialized,shangguan2024tomatoassessingvisualtemporal,burgess2025video,grauman2022ego4dworld3000hours,grauman2024egoexo4dunderstandingskilledhuman} and new large-scale datasets emerging for training \citep{grauman2022ego4dworld3000hours,grauman2024egoexo4dunderstandingskilledhuman,fu2025gigahands}.
On the methodological side, overcoming the VLM failure modes identified in this study will require progress in efficient video encoding \citep{choudhury2024dontlooktwicefaster}, accurate detection of small-scale objects and subtle movements \citep{wanghoang2024poodle}, monocular 3D understanding \citep{yang2024think}, and maintaining long-term spatial memory for occluded objects \citep{Plizzari2025OSNOM}.%Based on our results, we believe that achieving the level of human motion understanding needed for data-driven stroke rehabilitation and other video-based digital health applications will require new datasets and  methodological advances. %, we believe that new datasets and me

%% file: sections/04_method.tex
\section*{Methods}
\label{sec:method}

\subsection*{Vision-language models for stroke rehabilitation}

Vision-language models (VLMs) are powerful multimodal models that can function as video question-answering systems. Their inputs are a set of video frames and a free-form textual question or instruction called a ``prompt." Their output is a free-form textual answer. In simplified terms, a VLM consists of a pre-trained vision encoder connected to a large-language model (LLM) with a vision-to-text connector. The textual and visual inputs to a VLM are first converted to tokens, which are numerical vector representations that serve as the fundamental units for reasoning and generation. The tokens are then fed through the LLM to generate a sequence of output tokens. This sequence is converted to text to produce the final output of the VLM.\footnote{VLMs generate output tokens one at a time by sampling over a probability distribution over possible next tokens. For all experiments in this paper, we always select the most probable token.}

VLMs are trained in two stages \citep{liu2023visualinstructiontuning}. In the first stage, the connector learns to align the visual and textual representations using a large dataset of image-text pairs. The second stage fine-tunes the vision encoder, connector, and large-language model jointly to answer questions, describe scenes, and follow instructions about images and videos. In this work, we used $15$ state-of-the-art open-source vision-language models of various sizes from $6$ model families released in the past two years: LLaVA-NeXT-Video %(04/2024)
\citep{zhang2024llavanextvideo}, LLaVA-OneVision %(08/2024) 
\citep{li2024llava}, NVILA %(12/2024) 
\citep{liu2024nvila}, Qwen2.5-VL %(02/2025) 
\citep{bai2025qwen25vltechnicalreport}, InternVL3 %(04/2025) 
\citep{zhu2025internvl3exploringadvancedtraining}, and InternVL3.5 %(08/2025) 
\citep{wang2025internvl3}. More details on how inputs are processed for each model family are provided in App.~\ref{app:preprocessing}.\footnote{Our evaluations were done with \lstinline{lmms_eval} \citep{zhang2024lmmsevalrealitycheckevaluation,lmms_eval2024}.}

%Under the hood, this question-answering system operates in three main stages: pre-processing, inference (feed-forward through a neural network's layers), and post-processing. Since inference can only input and output \textit{tokens}---numerical vector representations that serve as the fundamental units for reasoning and generation---VLMs must wrap the inference step with a pre-processing and post-processing step to allow the system to answer text questions with text. More specific details regarding pre-processing and the pre-processing steps for each model are deferred to App.~\ref{app:preprocessing}.

As described in Figure~\ref{fig:fig1}, we formulated several problems relevant to stroke rehabilitation as human-motion recognition tasks that can be addressed using VLMs. For each task, image frames from rehabilitation videos and a task-defining prompt were provided as input to the VLM. In some cases, we optimized the prompt on held-out subjects to improve results and evaluated on a test set consisting of a disjoint set of patients. In the remainder of this section, we explain our formulation of each task, how we applied VLMs to solve it, the metrics used to evaluate performance, and any additional engineering aspects.

\subsection*{Activity identification}
\label{res:activity_id}

As shown in Figure~\ref{fig:fig1}(b), we formulated activity identification as a classification task, where the classes are nine rehabilitation activities. We provided the VLM with two inputs: eight frames, uniformly sampled from a rehabilitation video, and a classification prompt that describes the nine activities and asks the model to identify which one is depicted in the video frames. The textual output of the VLM indicates the estimated activity. The VLM used for this task is Qwen2.5-VL-7B-Instruct. We evaluated our approach on $640$ videos featuring $18$ healthy subjects and $51$ stroke patients (``(a) Activity Identification" on the right of Table~\ref{tab:stroke_cohort}).

We considered two alternative strategies for producing activity descriptions in the classification prompt. The first strategy was \emph{direct prompting}, where we generated one-sentence descriptions of the activities using ChatGPT 5 (\href{https://chat.openai.com}{https://chat.openai.com}) by providing screenshots of the supplementary material in \cite{kaku2022strokerehab}. The second strategy was \emph{optimized prompting}, specifically adapted to the Qwen2.5-VL-7B-Instruct model. We first asked the VLM for a description of the objects in the work station for held-out videos from two control subjects. We then modified the activity descriptions to match the wording of the model. For example, ``comb" became ``small rectangular object." Surprisingly, small changes like this boosted the accuracy from $53.4\%$ to $77.5\%$. We include transcripts of both prompts in App.~\ref{app:activity_prompts}.

\subsection*{Dose quantification}

As shown in Figure~\ref{fig:fig1}(c), we formulated the problem of measuring rehabilitation dosage as a sequence estimation task, where the goal is to identify the sequence of fine-grained motions carried out by a subject during a video. The five motions of interest, called \textit{functional primitives}, are \textit{reach} (move to contact a target object), \textit{reposition} (move proximate to a target object, e.g., the initial neutral spot), \textit{transport} (move a grasped target object), \textit{stabilize} (hold a target object still), and \textit{idle} (stand at the ready near a target object). In order to quantify rehabilitation dose from an estimated sequence of primitives, we simply count how many times each primitive occurs in the sequence.

To estimate the sequence of primitives, we first divided the video into segments with duration $0.533$ s. For each segment, the VLM received two inputs: eight frames uniformly sampled from the segment and a prompt requesting primitive identification. The VLM text output identified the estimated functional primitive for each segment, and the resulting outputs were concatenated to form the estimated sequence for the video. We constrained the model to one primitive per segment because initial experiments allowing multiple primitives per segment led to overcounting. We report results for alternative segment durations and frame sampling rates using Qwen2.5-VL-32B-Instruct in App.~\ref{app:further_results_prims_id}.

%Initially, we prompted the VLM to output \textit{all} primitives in a segment. However, this yielded near-identical outputs and excessive over-segmentation. As such, we proceeded by 
In each segment, primitive classification was carried out based on the insight that classification of motion and grasp essentially suffice to identify primitives, as described in Table~\ref{tab:prim_def}. The motion and grasp information was obtained via \emph{Decomposed Prompting}, feeding the following two binary questions to the VLM: (1) Is the hand moving significantly? and (2) Is the hand grasping an object? %Table~\ref{tab:prim_def} illustrates how the answers to these questions can be used to reconstruct the segment primitive. 
The answers to these questions were complemented with a heuristic to distinguish \textit{reach} and \textit{reposition} primitives % yield the same answers to these questions, we needed a method to distinguish them 
based on the presence of a terminal grasp. %We applied a simple rule: i
If the VLM predicts either \textit{transport} or \textit{stabilize} within two seconds of the primitive's start time, we classify the primitive as \textit{reach;} otherwise, it is classified as \textit{reposition.} The 2-second threshold was validated on $21$ videos from two control subjects outside the evaluation set. Further evidence supporting this rule's efficacy is provided by the strong performance of the \textit{Omniscient} baseline, described next.
% This threshold was set based on the ground truth sequences for two control subjects not in the evaluation set ($21$ videos). We kept the other primitives the same and pretended we needed to classify between \textit{reach} and \textit{reposition} in their original positions. The $2$-s threshold perfectly reconstructed the sequences for these videos.

\begin{table}[t]
\centering
\caption{\textbf{Definition and motion-grasp decomposition of functional primitives for dose quantification.} The table provides definitions of the five functional primitives~\cite{schambra2019taxonomy}, and shows how answers to the questions ``Is the hand moving significantly?" and ``Is the hand grasping an object?" can be used to identify a primitive (except for reach and reposition, which must be distinguished based on terminal grasp).}
\begin{tabular}{lllll}
Primitive & Definition & Motion & Grasp & Grasp at end \\
\midrule
Reach       & Move into contact with a target object & Present & No & Yes \\
Transport   & Convey a target object & Present & Yes & Either \\
Reposition  & Move without the purpose of future contact & Present & No & No \\
Stabilize   & Keep a target object still & Minimal & Yes & Either \\
Idle        & Stand at the ready & Minimal & No & Either \\
\bottomrule
\end{tabular}
\label{tab:prim_def}
\end{table}

%Having explained how the VLMs were prompted to generate Table~\ref{tab:smc_llms}, 
We designed two baselines to provide context for the results in Table~\ref{tab:smc_llms}. The \textit{Omniscient} baseline represents the optimal performance achievable by a VLM operating under the constraints of producing only one primitive per segment and applying the $2$-second heuristic to distinguish \emph{reach} and \emph{reposition}.
For each video segment, the ground-truth primitive was considered to be the one at the midpoint of the video, from which the ground-truth \textit{motion} and \textit{grasp} states were derived. 
The estimated sequence was then reconstructed as described in the previous paragraph. 
The \textit{Markov} baseline is a baseline that has access to first-order transition statistics of the ground-truth sequences, but not to the video itself. We used the $90$-video test set (see ``(b) Dose Quantification" on the right of Table~\ref{tab:stroke_cohort}) to estimate the transition probabilities for grasp and motion. 
%two $2$x$2$ probability transition matrices (one for motion, one for grasp) based on transitions at the segment level. 
We then generated artificial predictions by starting in the no-motion/no-grasp state and randomly sampling transitions for subsequent segments. 
The primitive sequence was reconstructed from the motion/grasp sequence as in the previous paragraph.

\subsection*{Optimizations for dose quantification}

Here, we describe various optimizations to the \textit{Decomposed Prompting} strategy from the previous subsection. We evaluated these optimizations using Qwen2.5-VL-32B-Instruct as the VLM given its solid performance and inference speed.

\textbf{Contextual prompting:} In an attempt to reduce over-segmentation (see Fig.~\ref{fig:vlm_analysis}(c)), we incorporated the VLM's predictions from the previous segment directly into the text prompt for the current segment (see App.~\ref{app:prim_id_prompts} for the prompts).

\textbf{Cropping:} Since most primitives can be identified by observing the hand in the upper extremity of interest, we employed a 2D pose model \citep{Jocher_Ultralytics_YOLO_2023} to localize and crop the hand region, producing cropped images that were provided as visual input to the VLM. First, we prompted the VLM to identify all individuals in the initial video frame and designated the person with the largest bounding box as the subject. %Second, we localized the hand within every frame of each segment. 
We then used a 2D pose model to obtain COCO keypoints \citep{lin2014microsoft} for the subject. We selected the elbow and wrist keypoints %associated with the subject and extract the elbow and wrist keypoints with associated confidences. Afterwards, we 
and estimated the hand region by extending the elbow-to-wrist vector by a factor of $0.7$. %Finally, we attempted to crop for each segment individually, with associated crop signals. 
If, for any frame in a segment, the lowest confidence score among the elbow and wrist keypoints was below $0.9$, we abstained from cropping and passed on the original frames to the VLM. If quick movement was detected, we extracted a moving crop that linearly interpolated the hand position across the segment; otherwise, we extracted a still crop centered at the hand's average position over the segment. Quick movement was detected when the average pixel movement in the horizontal or vertical direction exceeded $15$. We ensured that crops respected image boundaries. The crop size was $224$x$224$ pixels.

\textbf{Pose-Refined promptIng Module (PRIM-RS):} We propose PRIM-RS as a pipeline optimized for dose quantification for the RTT and shelf tasks (results shown in Fig.~\ref{fig:rtt_shelf_counts}). This system, which was tuned on a held-out set of videos from two control subjects and one severe patient, employs specialized binary prompts, a pose-informed decision pathway, and carefully designed post-processing.

\underline{Optimized Prompting:} We adapted the two binary prompts used in \textit{Decomposed Prompting} as follows. First, we replaced motion detection with idle detection. Rather than asking whether the hand is in motion, the prompt queried whether the hand is still and not interacting with any object. This change was motivated by the observation that the VLM reliably detects idle hands, particularly during the repetitive act-and-rest cycles characteristic of the RTT and shelf tasks, and it generally performed well in practice.
Second, we modified grasp detection by introducing contextual prompting: the prompt asked whether the subject was picking up an object if the hand had been empty in the prior state, or releasing an object if one had been previously held. Contextual prompting yielded mixed results, and careful post-processing proved more critical for robust performance. Full prompt details are provided in App.~\ref{app:prim_id_prompts}.

\underline{Pose-Informed Decision Making:} PRIM-RS utilized a pose model for hand-region cropping, following the procedure detailed in the Cropping subsection. State assignment for each segment was handled as follows: if the cropping system abstained, the idle state was set to ``idle" and the grasp state to ``empty." If cropping succeeded and quick movement was detected, the idle state was set to ``not idle" and the grasp state was retained from the previous segment, avoiding potentially inaccurate VLM predictions during rapid motion. Otherwise, we prompted the VLM to estimate both idle and grasp states. Concatenating these states resulted in an idle and grasp signal for the video.

\underline{Post-Processing:} Post-processing was executed by sequentially filling an empty list with a length equal to the number of segments in the video, addressing the primitives in the following order: \textit{idle}, \textit{reach} and \textit{reposition}, and \textit{transport} and \textit{stabilize.}

\begin{itemize}
    \item \emph{Idle:} Segments for the idle signal were relabeled in two passes: first to ``not idle" if adjacent segments were ``not idle," and then to ``idle" if adjacent segments were ``idle," using the updated labels from the first pass. Using the resulting smoothed idle signal, we corrected the grasp signal by setting any segment to ``empty" whenever the corresponding idle state was ``idle." We then applied the same smoothing operation to this updated grasp signal. Because such smoothing can mask short-duration events, we reduced the segment length from 0.533 s to 0.267 s (using four frames instead of eight) while maintaining the same frame rate.
    \item \emph{Reach and reposition:} We labeled \textit{reach} and \textit{reposition} for all ``not idle" and ``empty-handed” segments. Consecutive segments were grouped into contiguous blocks, and each block was classified based on whether its surrounding segments were ``idle" or ``holding," yielding four possibilities. Following Table~\ref{tab:prim_def}, each case was assigned one of \textit{reach}, \textit{reposition}, \textit{reach-reposition}, or \textit{reposition-reach.} Further, because direct idle–grasp or grasp–idle transitions were rare, we explicitly inserted \textit{reach} or \textit{reposition} where appropriate.
    \item \emph{Transport and stabilize:} Finally, \textit{transport} and \textit{stabilize} were assigned within ``holding" blocks. We detected stillness similarly to how quick movement was detected for cropping, but used $\leq3$ px instead of $\geq15$ px as the threshold. If, for a given block, stillness was detected for at least three consecutive segments, indicating confidence by the pose model, those three segments would be labeled as \textit{stabilize}. Further, if stillness was detected within three segments of the block's end, that segment and the following segments within the block were labeled as \textit{stabilize}. Other ``holding" segments defaulted to \textit{transport.}
\end{itemize}

\subsection*{Metrics for assessing dose quantification}

% Because primitives often meld into each other, we adopt sequence-level metrics on the \textit{de-duplicated} primitive sequences. For example, \lstinline|idle, idle, reach, transport, transport| becomes \lstinline|idle, reach, transport|. De-duplication drastically reduces the ground-truth sequence length since there may be dozens of frames with the same primitive label.

We employed three metrics to measure the quality of the estimated functional-primitive sequences for rehabilitation dose quantification. We denote the predicted sequence by $P$ and the ground-truth sequence by $G$. The \textit{edit score} $\text{ES}(G,P)$ and the \textit{action error rate} $\text{AER}(G,P)$ \citep{kaku2022strokerehab} are normalized versions of the \textit{Levenshtein edit distance} $\text{L}(G,P)$, which counts the minimum number of operations needed to transform one sequence into the other. The operations are insertion, deletion, and substitution. For example, \lstinline|idle, reach, transport| has an edit distance of $2$ from \lstinline|reach, stabilize, transport| (replace \textit{idle} with \textit{reach} and \textit{reach} with \textit{stabilize}, or delete \textit{idle} and add \textit{stabilize}).

\begin{equation*}
\begin{minipage}{0.6\linewidth}
\centering
$\text{ES}(G,P) = \left( 1 - \dfrac{L(G,P)}{\max(\text{len}(G), \text{len}(P))} \right) \times 100$
\end{minipage}\hfill
\begin{minipage}{0.36\linewidth}
\centering
$\text{AER}(G,P) = \dfrac{L(G,P)}{\text{len}(G)}$,
\end{minipage}
\end{equation*}
where len indicates the length of a sequence.

%The sensible ranges for ES ($\uparrow$) and AER ($\downarrow$) are $0$ to $100$ and $0$ to $1$, respectively. 
Note that a higher ES is better, while a lower AER is better.
For a null prediction, the ES is $0$ and the AER is $1$. We report both metrics because the ES is common in the action recognition literature \citep{farha2019ms,wang2020boundary,ishikawa2021alleviating,lei2018temporal,kaku2022strokerehab} and the AER is better suited to evaluating estimation of highly granular motions like functional primitives, as it
more heavily penalizes long incorrectly estimated sequences \cite{kaku2022strokerehab}.

The final metric is the \textit{relative counting error} $\text{RCE}(G,P)$, which directly evaluates rehabilitation dose quantification based on primitive counts. The RCE aggregates the counting errors across the primitives and normalizes the sum by the ground-truth sequence length:
\[
\text{RCE}(G,P)=\frac{1}{\text{len}(G)} \sum_{p\in\mathcal{P}} |c(p,G)-c(p,P)|,
\]
where $\mathcal{P}$ denotes the set containing the five primitives and $c(p,S)$ the number of times primitive $p$ occurs in the sequence $S$.

\subsection*{Impairment quantification}

\begin{figure}
    \centering
    \includegraphics[width=\linewidth]{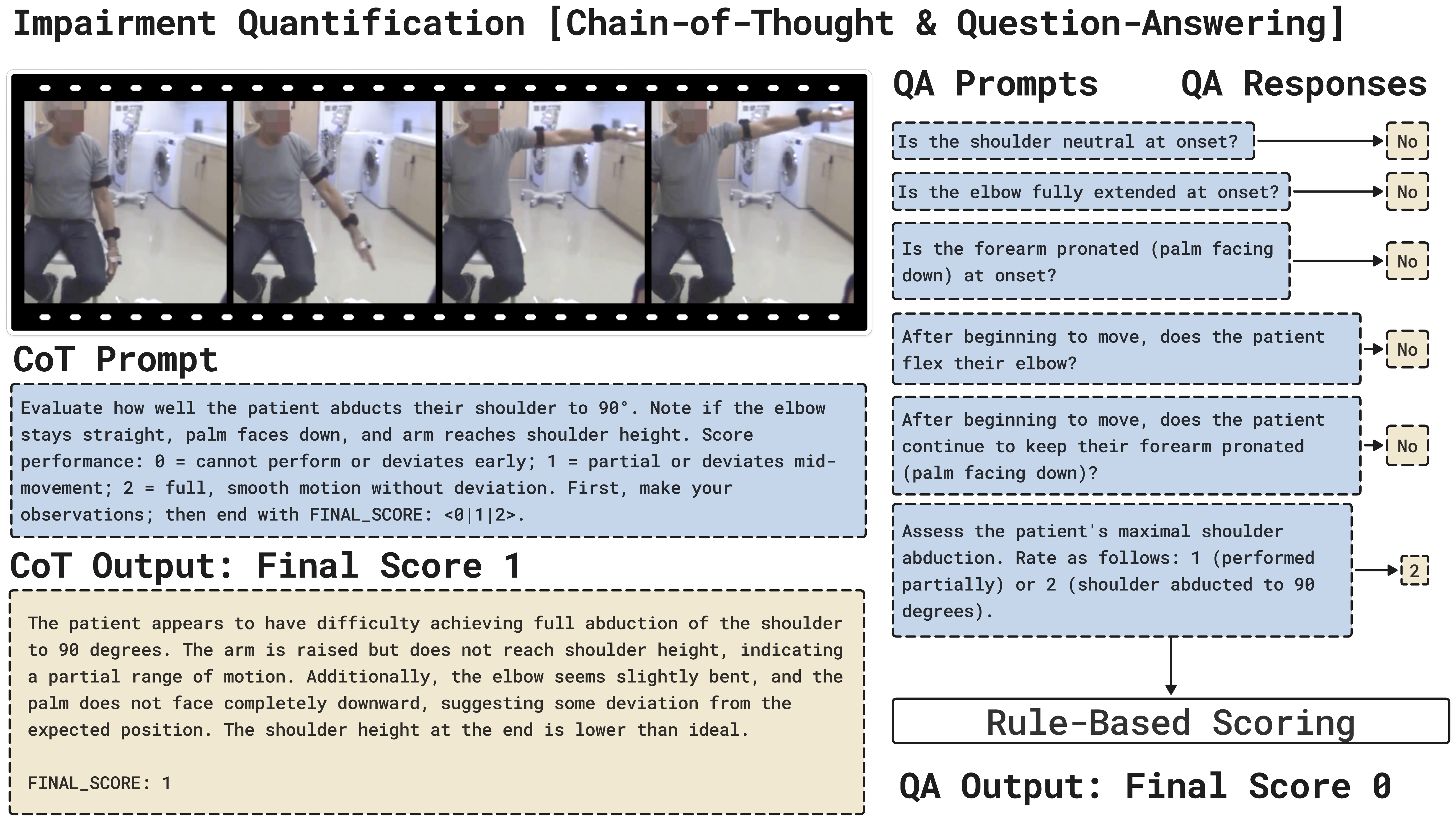}
    \caption{\textbf{Strategies for Applying VLMs to Impairment Quantification.} To assess impairment using VLMs, we evaluated two approaches. Chain-of-Thought supplies the VLM with all relevant contextual information in the prompt, requesting the model to reason through its observations and conclude with a final score. Rule-based question-answering prompts the VLM with a list of binary questions. The answers are then aggregated to deduce a final score. Here, the first ``No" set the final score to ``0" (the rest of the answers do not matter, but are shown for illustration). For this specific video, both methods ended up with an incorrect score: the ground-truth score was $2$.}
    \label{fig:impairment_quantification}
\end{figure}

As shown in Figure~\ref{fig:fig1}(d), we leveraged the Fugl-Meyer Assessment (FMA) to perform impairment quantification using VLMs. The FMA consists of multiple items designed to evaluate the movement quality of a subject. We used the VLM to rate each item by providing  (1) a short clip of a subject performing an FMA item and (2) the rating instructions given to a trained expert for that item. The textual output of the VLM was then used to estimate the score for the item: $0$ (worst), $1$, or $2$ (best). 

We employed two general strategies for prompting, explained in Fig.~\ref{fig:impairment_quantification}. For \emph{Chain-of-Thought (CoT)} prompting \citep{wei2023chainofthoughtpromptingelicitsreasoning}, we provided the VLM with all relevant rating information in the prompt and asked it to reason through its observations and conclude with a final score. For \emph{Rule-based Question-Answering (QA),} we designed a list of questions tailored to each FMA item. Each question elicited a binary response. Depending on the response, a score was assigned or the next question was presented. This process continued until the final question, which forced a final rating. See App.~\ref{app:impairment_quant_prompts} for the specific COT and QA FMA prompts.

%We now elucidate a few more final details regarding the visual input. 
Since the subject performed each FMA item three times, we selected the middle repetition for the VLM to rate, using the start and end times of this middle repetition to define clip boundaries. Each subject was filmed using two camera angles throughout the FMA session: one positioned directly in front of the subject and another facing their paretic side (see App.~\ref{app:fma_videos}). We chose the camera angle best suited for rating a particular question.

For most FMA items, we uniformly selected eight frames from the clip. This low number should be sufficient for accurate rating because the sampling rate captures the relevant motion characteristics and the clips are short ($95\%$ have duration $<10$ s; see App.~\ref{app:fma_videos}). 
An exception is the FMA section named \emph{Coordination/Speed}, where subjects are instructed to rapidly move their finger between their nose and knee five times to test for tremor, dysmetria, and movement speed. For tremor and dysmetria, we segmented the video using eight frames per $0.267$-s segment, rated each segment, and averaged the rating across segments to generate a final rating. For movement speed, we similarly segmented the video and prompted the VLM to count the number of ``touches" detected on the nose and knee within each segment. We then identified the minimum time point at which the cumulative touch counts for both knee and nose exceeded five. The final score was then determined by comparing this minimum time point between the paretic and healthy sides.

%% file: sections/05_appendix.tex
\begin{appendices}

\newpage
\begin{center}
\huge
\textbf{Supplementary Material}
\end{center}

\section*{Contents}

\begin{itemize}
  \item \hyperref[app:preprocessing]{VLM Input Preprocessing}
  \item \hyperref[app:activity_videos]{More Details on the Activity Videos}
  \item \hyperref[app:activity_prompts]{Prompts for Activity Identification}
  \item \hyperref[app:prim_id_prompts]{Prompts for Primitives Identification}
  \item \hyperref[app:further_results_prims_id]{Further Results for Primitives Identification}
  \item \hyperref[app:left_v_right]{Figure~\ref{fig:vlm_analysis}(d) Experimental Procedure}
  \item \hyperref[app:fma_videos]{More Details on the Fugl-Meyer Assessment Videos}
  \item \hyperref[app:impairment_quant_prompts]{Prompts for Impairment Quantification}
\end{itemize}

\section{VLM Input Preprocessing}
\label{app:preprocessing}

We performed our evaluations with greedy decoding (temperature of $0$, top-p set to None, and a beam size of $1$). Table~\ref{tab:model_overview} shows hyperparameter configurations for the model families.%, with a note on Qwen2.5-VL pre-processing.

%Pre-processing must be done on visual and text input. To pre-process visual input, VLMs transform the image (e.g. normalization, resizing), divide the image into small square patches, and feed-forward the patches through a fine-tuned image encoder to obtain visual tokens. To pre-process text input, VLMs break up the text into segments and perform a look-up in a dictionary that converts each segment into a text token. After obtaining these tokens, the pre-processing step may also include modifying them with positional embeddings (e.g. extremely important for helping the neural network identify the spatial and temporal position of the patch a visual token corresponds to, or to establish word ordering in the text tokens) and adding formatting tokens.

\begin{table}[ht]
\centering
\caption{\textbf{Model Configurations.} We used the default model configurations provided by \lstinline{lmms_eval} and list important settings here.}
\label{tab:model_overview}
% Using 'l' for the first column and 'X' columns (from tabularx)
% for the second and third to automatically wrap text and fit the page width.
% >{\raggedright\arraybackslash} ensures the wrapped text is left-aligned.
\begin{tabularx}{\textwidth}{
  >{\raggedright\arraybackslash\hsize=0.15\hsize}X  % First column (narrower)
  >{\raggedright\arraybackslash\hsize=0.25\hsize}X  % Middle column (default)
  >{\raggedright\arraybackslash\hsize=0.5\hsize}X  % Last column (wider)
}
\toprule
\textbf{Model Family} & \textbf{Configuration / Settings} & \textbf{Model Tags (Hugging Face Hub)} \\
\midrule
LLaVA-NeXT-Video \citep{zhang2024llavanextvideo} & conv\_template=qwen\_1\_5\textsuperscript{a} & \begin{tabular}[t]{@{}l@{}}lmms-lab/LLaVA-NeXT-Video-7B-Qwen2 \\ lmms-lab/LLaVA-NeXT-Video-72B-Qwen2\end{tabular} \\
\addlinespace % Adds a small vertical space

LLaVA-OneVision \citep{li2024llava} & \begin{tabular}[t]{@{}l@{}}conv\_template=qwen\_1\_5\textsuperscript{a} \\ model\_name=llava\_qwen\end{tabular} & \begin{tabular}[t]{@{}l@{}}lmms-lab/llava-onevision-qwen2-0.5b-ov \\ lmms-lab/llava-onevision-qwen2-7b-ov \\ lmms-lab/llava-onevision-qwen2-72b-ov-sft\end{tabular} \\
\addlinespace

NVILA \citep{liu2024nvila} & conv\_template=auto\textsuperscript{a} & \begin{tabular}[t]{@{}l@{}}Efficient-Large-Model/NVILA-8B \\ Efficient-Large-Model/NVILA-15B\end{tabular} \\
\addlinespace

Qwen2.5-VL \citep{bai2025qwen25vltechnicalreport} & \begin{tabular}[t]{@{}l@{}} qwen-vl-utils: v0.0.11\textsuperscript{b} \\ min\_pixels: 256*28*28 \\ max\_pixels: 1,605,632\end{tabular} & \begin{tabular}[t]{@{}l@{}}Qwen/Qwen2.5-VL-7B-Instruct \\ Qwen/Qwen2.5-VL-32B-Instruct \\ Qwen/Qwen2.5-VL-72B-Instruct\end{tabular} \\
\addlinespace

InternVL3 / 3.5 \citep{zhu2025internvl3exploringadvancedtraining, wang2025internvl3} & \begin{tabular}[t]{@{}l@{}}modality=video \\ input\_size=448\textsuperscript{c} \\ max\_num=1\textsuperscript{d} \\ use\_thumbnail=True\textsuperscript{e}\end{tabular} & \begin{tabular}[t]{@{}l@{}}OpenGVLab/InternVL3-78B \\ OpenGVLab/InternVL3\_5-2B \\ OpenGVLab/InternVL3\_5-8B \\ OpenGVLab/InternVL3\_5-38B \\ OpenGVLab/InternVL3\_5-30B-A3B\end{tabular} \\
\bottomrule
\end{tabularx}
% Footnotes section
\begin{flushleft} \small
\textsuperscript{a} \textbf{conv\_template}: Conversation template. \\
\textsuperscript{b} \textbf{input\_size}: The pre-processing library, \lstinline{qwen-vl-utils} was updated in version 0.0.14 to resize based on \lstinline{min_pixels} and \lstinline{max_pixels} (unlike version 0.0.11, used in this study). \\
\textsuperscript{c} \textbf{input\_size}: The resolution (448x448) to which each image patch is resized. \\
\textsuperscript{d} \textbf{max\_num}: The maximum number of patches to split each frame into. \\
\textsuperscript{e} \textbf{use\_thumbnail}: Whether to include a downsampled thumbnail of the entire frame as an additional patch.
\end{flushleft}
\end{table}

\paragraph*{Qwen2.5-VL Pre-processing}

The pre-processing pipeline for Qwen2.5-VL does very minor reshaping. When given a video of eight frames at $704 \times 1088$ resolution, it proceeds as follows:

\begin{enumerate}
    \item \textbf{Divisibility Resizing:} The frames are first resized to $\mathbf{700 \times 1092}$. This is a minor adjustment to ensure both the height ($700$) and width ($1092$) are divisible by $28$. This factor is derived from the spatial patch size ($p=14$) and temporal patch size ($\tau=2$).

    \item \textbf{Pixel Constraint Check:} The \lstinline{min_pixels} and \lstinline{max_pixels} configurations in Table~\ref{tab:model_overview} are per-frame limits. The new resolution ($700 \times 1092 = 764,400$ pixels) is comfortably within this range, so no further resizing is needed to meet these constraints.

    \item \textbf{Patching and Tokenization:} The video tensor, now with shape $(T, C, H, W) = (8, 3, 700, 1092)$, is converted into patch tokens.
    \begin{itemize}
        \item The $T=8$ frames are grouped temporally by $\tau=2$, resulting in $\mathbf{4}$ \textbf{temporal groups}.
        \item Each frame is divided spatially into $p \times p = 14 \times 14$ patches.
        \item This yields $(H/p) \times (W/p) = (700/14) \times (1092/14) = 50 \times 78 = \mathbf{3,900}$ \textbf{spatial patches} per temporal group.
        \item The total number of initial patches (or "tubelets") sent to the visual encoder is $4 \text{ (groups)} \times 3,900 \text{ (patches/group)} = \mathbf{15,600}$.
    \end{itemize}

    \item \textbf{Token Merging:} After the visual encoder, a patch merger module reduces the number of tokens by a factor of $4$.

    \item \textbf{Final Token Count:} The total number of pure vision tokens produced by this process is 15,600 / 4 = \textbf{3,900} tokens. These tokens are fed into the LLM, along with the prompt tokens.
\end{enumerate}

% InternVL3

% InternVL3.5

\newpage
\section{More Details on the Activity Videos}
\label{app:activity_videos}

\begin{figure}[h!]
    \centering
    \includegraphics[width=\linewidth]{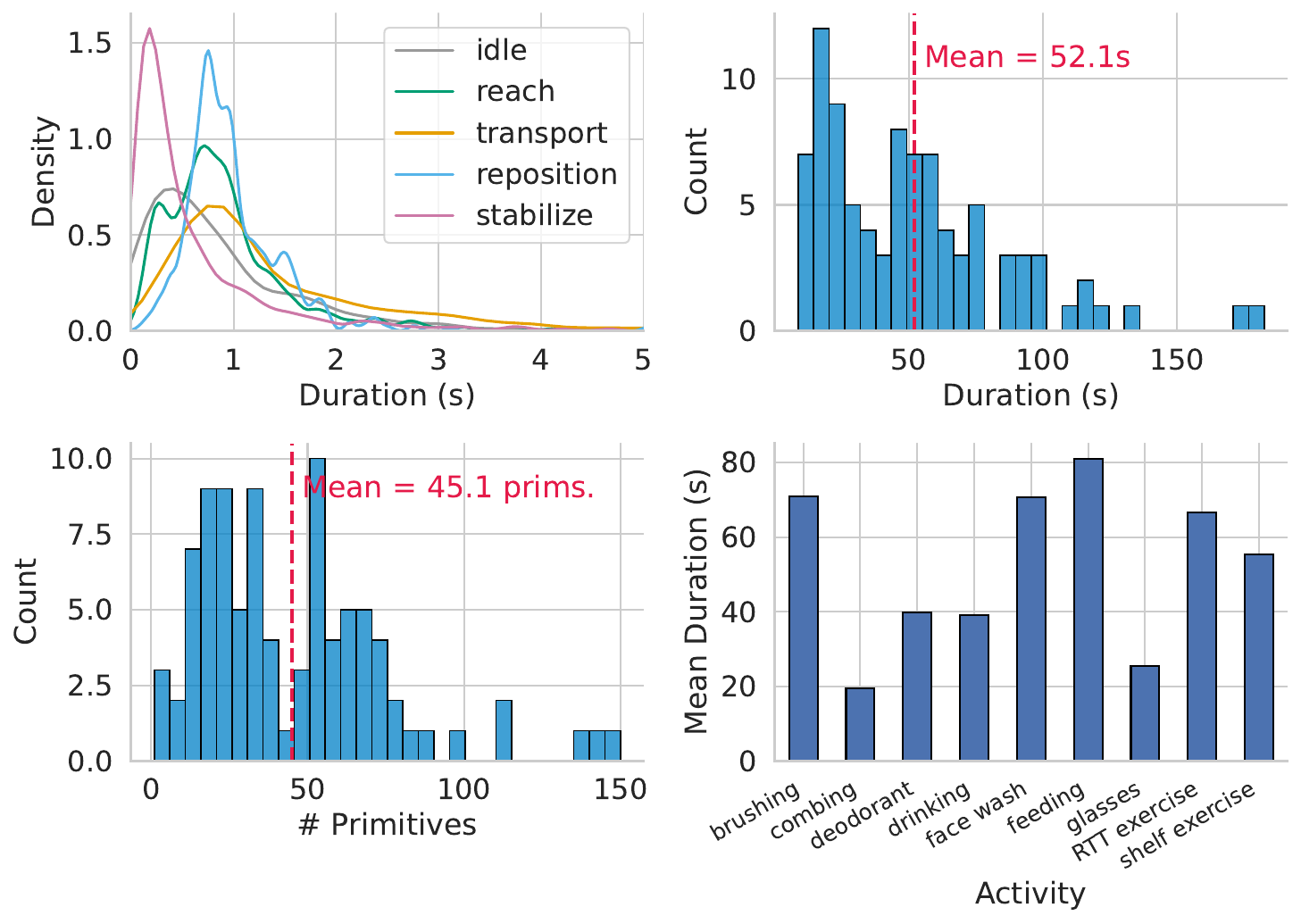}
    \caption{\textbf{Dataset statistics for the $90$-video data set used for primitives identification evaluation.} \textbf{Top left:} Kernel density estimate of primitive durations across all videos. \textbf{Top right:} Histogram of the video durations. \textbf{Bottom left:} Histogram of the number of primitives per video. \textbf{Bottom right:} Mean video durations across the nine activities.}
    \label{fig:activity_metadata_plot}
\end{figure}

Here, we provide visualizations into dataset statistics, as well as further details on each activity. See Fig.~\ref{fig:activity_metadata_plot} and Table~\ref{tab:act_descp_1}.

\begin{table*}[!ht]
\caption{Description of the activities performed by the stroke-impaired patients in the cohort. Sourced from \citep{kaku2022strokerehab}.}
\resizebox{\textwidth}{!}{%
\centering
\begin{tabular}{|p{2cm}|p{5cm}|p{2cm}|p{5cm}|}
\hline
\multicolumn{1}{|c|}{Activity} & \multicolumn{1}{c|}{Workspace} & \multicolumn{1}{c|}{\begin{tabular}[c]{@{}c@{}}Target\\ object(s)\end{tabular}} & \multicolumn{1}{c|}{Instructions} \\ 
% \cline{4-5} 
% \multicolumn{1}{|c|}{} & \multicolumn{1}{c|}{} & \multicolumn{1}{c|}{} & \multicolumn{1}{c|}{Proximal \textgreater Distal} & \multicolumn{1}{c|}{Proximal \textless Distal} \\ 
\hline
% Activity & Workspace & Target object(s) & Instructions \\ \hline
Washing face & Sink with a small tub (32.3 x 24.1 x 2.5 cm³) in it and two folded washcloths on either side of the countertop, 30 cm from edge closest to patient & Washcloths, faucet handle, and tub & Fill tub with water, dip washcloth on the right side into water, wring it, wiping each side of their face with wet washcloth, place it back on countertop. Use washcloth on the left side to dry face, place it back on countertop \\ \hline
Applying deodorant & Tabletop with deodorant placed at midline, 25 cm from edge closest to patient & Deodorant (solid twist-base) & Remove cap, twist base a few times, apply deodorant, replace cap, untwist the base, put deodorant on table \\ \hline
Hair combing & Tabletop with comb placed at midline, 25 cm from edge closest to patient & Comb & Pick up comb and comb both sides of head \\ \hline
Don/doffing glasses & Tabletop with glasses placed at midline, 25 cm from edge closest to patient & Pair of glasses & Wear glasses, return hands to table, remove glasses and place on table\\ \hline
Eating & Table top with a standard-size paper plate (21.6 cm diameter) placed at midline, 2 cm from edge, utensils placed 3 cm from edge, 5 cm from either side of plate, a baggie with a slice of bread placed 25 cm from edge, 23 cm left of midline, and a margarine packet placed 32 cm from edge, 17 cm right of midline & Paper plate, fork, knife, re-sealable sandwich baggie, slice of bread, single-serve margarine container & Remove bread from plastic bag and put it on plate, open margarine pack and spread it on bread, cut bread into four pieces, cut off and eat a small bite-sized piece \\ \hline

Drinking & Tabletop with water bottle and paper cup 18 cm to the left and right of midline, 25 cm from edge closest to patient & Water bottle (12 oz),
paper cup (4 oz) & Open water bottle, pour water into cup, take a sip of water, place cup on table, and replace cap on bottle \\ \hline
Tooth brushing & Sink with toothpaste and toothbrush on either side of the countertop, 30 cm from edge closest to patient & Travel-sized toothpaste, toothbrush with built-up foam grip, faucet handle & Wet toothbrush, apply toothpaste to toothbrush, replace cap on toothpaste tube, brush teeth, rinse toothbrush and mouth, place toothbrush back on countertop \\ \hline
Moving object on a horizontal surface & Horizontal circular array (48.5 cm diameter) of 8  targets (5 cm diameter) & Toilet paper roll wrapped in self-adhesive wrap  & Move the roll between the center and each outer target, resting between each motion and at the end \\ \hline
Moving object on/off a Shelf & Shelf with two levels (33 cm and 53 cm) with 3 targets on both levels (22.5 cm, 45 cm, and 67.5 cm away from the left-most edge) & Toilet paper roll wrapped in self-adhesive wrap  & Move the roll between the center target and each target on the shelf, resting between each motion and at the end \\ \hline
\end{tabular}}

\label{tab:act_descp_1}
\end{table*}

\newpage
\section{Prompts for Activity Identification}\label{app:activity_prompts}

Listed below are the direct and tailored prompts for activity identification, respectively.

\begin{lstlisting}[caption={Activity Identification Direct Prompt}]
{
  "question": "Which activity is the patient performing in this video?",
  "response_instructions": "After noting your observations, end your reply with exactly one line: FINAL_ANSWER: <activity_name>",
  "instructions": "Determine the main activity being performed by the patient. Choose the most fitting activity label from the list below. Always respond with an activity, even if uncertain.",
  "activity_classes": {
    "Brushing": "The patient applies toothpaste to a toothbrush, brushes their teeth, rinses, and sets the brush back down.",
    "Combing": "The patient picks up a comb and combs both sides of their hair.",
    "Deodorant": "The patient twists open a deodorant stick, applies it under the arm, then replaces the cap.",
    "Drinking": "The patient pours water from a bottle into a cup, takes a sip, and replaces the cap.",
    "Face wash": "The patient washes and dries their face using two washcloths at a sink.",
    "Feeding": "The patient prepares bread with margarine on a plate and eats a small piece using utensils.",
    "Glasses": "The patient puts on or removes a pair of glasses from the tabletop.",
    "RTT exercise": "The patient slides a toilet paper roll between center and outer targets on a flat surface.",
    "Shelf exercise": "The patient transfers a toilet paper roll between the center target and multiple shelf levels."
  }
}
\end{lstlisting}

\begin{lstlisting}[caption={Activity Identification Tailored Prompt}]
{
  "question": "Which activity is the patient performing in this video?",
  "response_instructions": "After noting your observations, end your reply with exactly one line: FINAL_ANSWER: <activity_name>",
  "instructions": "Determine the main activity being performed by the patient. Choose the most fitting activity label from the list below. Always respond with an activity, even if uncertain.",
  "activity_classes": {
    "Brushing": "The patient is at the sink. Either toothpaste or a toothbrush is visible, indicating the patient is brushing their teeth.",
    "Combing": "The patient grabs a small rectangular object (likely a comb) on the table and moves it near the hair area (likely to groom their hair).",
    "Deodorant": "The patient applies deodorant using a deodorant tube (likely white) on the table. The hand is seen moving towards the underarm area.",
    "Drinking": "The patient pours water from a plastic water bottle into a cylindrical cup on the table and takes a sip.",
    "Face wash": "The patient washes their face at the sink using water and a wash cloth or towel.",
    "Feeding": "The patient prepares and eats bread on a white plate by spreading margarine, cutting it, and taking bites.",
    "Glasses": "The patient grabs a pair of glasses from the table and puts them on or removes them.",
    "RTT exercise": "The patient repeatedly moves a cylindrical block on the table. ",
    "Shelf exercise": "The patient repeatedly moves a cylindrical block on a TRANSPARENT shelf.",
  }
}
\end{lstlisting}

\newpage
\section{Prompts for Primitive Identification}\label{app:prim_id_prompts}

Here, we list the prompts used in our experiments for primitive identification. We first tested an \textit{Ideal Primitives Identification Prompt} that asks for a sequence of primitives per segment, whose poor performance led to the \textit{Single-Prediction Primitives Identification Prompt}. Afterwards, we tested \textit{Decomposed Prompting} and a \textit{Contextual Prompting}. Finally, refer to \textit{PRIM-RS Prompts} for the RTT/shelf-specific pipeline.

For each prompt, we chose the text within the parentheses, separated by ``|", to cater to the specific situation. For contextual prompting, we additionally chose the hand reference (e.g. ``hand in the center" vs. ``patient's LEFT hand") based on whether cropping was successful.

\begin{lstlisting}[caption={Ideal Primitives Identification Prompt}]
Focus on the patient's (LEFT|RIGHT) hand. Output the sequence of functional primitives performed by the patient's (LEFT|RIGHT) hand as a comma-separated list.

Functional primitives:
- IDLE: hand is waiting
- REACH: hand in motion with the purpose of contact with an object
- REPOSITION: hand in motion with no contact at the endpoint
- STABILIZE: hand steady to keep a target object still
- TRANSPORT: hand in motion to convey an object in space
Only output the functional primitives (no definitions) as a comma-separated list.
\end{lstlisting}

\begin{lstlisting}[caption={Single-Prediction Primitives Identification Prompt}, label={lst:sp}]
Focus on the patient's (LEFT|RIGHT) hand. Output the functional primitive performed by the patient's (LEFT|RIGHT) hand as a single word.

Functional primitives:
- IDLE: hand is waiting
- REACH: hand in motion with the purpose of contact with an object
- REPOSITION: hand in motion with no contact at the endpoint
- STABILIZE: hand steady to keep a target object still
- TRANSPORT: hand in motion to convey an object in space
Only output one functional primitive.
\end{lstlisting}

\begin{lstlisting}[caption={Decomposed Prompting (Motion) (1/2)}, label={lst:dpm}]
Focus on the patient's (LEFT|RIGHT) hand. Is it actively moving an object, moving towards an object, or moving away from an object? Answer YES or NO.
\end{lstlisting}

\begin{lstlisting}[caption={Decomposed Prompting (Grasp) (2/2)}, label={lst:dpg}]
Focus on the patient's (LEFT|RIGHT) hand. Is it actively grasping or holding an object? Answer YES or NO.
\end{lstlisting}

\begin{lstlisting}[caption={Contextual Prompting (Motion) (1/2)}]
Focus on the (hand in the center|patient's LEFT hand|patient's RIGHT hand). It was previously (still|moving an object or moving toward/away from one). Is it now (actively moving an object, moving towards an object, or moving away from an object|still)? Answer YES or NO.
\end{lstlisting}

\begin{lstlisting}[caption={Contextual Prompting (Grasp) (2/2)}]
Focus on the (hand in the center|patient's LEFT hand|patient's RIGHT hand). Previously, (it was actively grasping an object|the hand was empty). Does it (release the object|grasp an object) in this clip? Answer YES or NO directly.
\end{lstlisting}

\begin{lstlisting}[caption={PRIM-RS Prompts (Idle) (1/3)}]
Is (the hand|the patient's LEFT hand|the patient's RIGHT hand) idle in this video clip?
(Idle) Visibly resting on the black mat, not moving, and not interacting with a cylindrical block. (Active) In the air, moving towards an object, moving away from an object, interacting with a cylindrical block, or 'resting' on a cylindrical block. The hand can be moving very slowly through the air and still be considered 'active.' Answer 'Yes.' if (the hand|the patient's LEFT hand|the patient's RIGHT hand) is idle; answer 'No.' otherwise.
\end{lstlisting}

\begin{lstlisting}[caption={PRIM-RS Prompts (Grasp) (2/3)}]
This is a chunk from a video sequence. In the previous chunk, (the hand|the patient's LEFT hand|the patient's RIGHT hand) was not holding anything. Answer directly: 'Yes.' if (the hand|the patient's LEFT hand|the patient's RIGHT hand) visibly picks up a cylindrical block in this chunk; answer 'No.' otherwise.
Mere contact does not count as grasping.
\end{lstlisting}

\begin{lstlisting}[caption={PRIM-RS Prompts (Release) (3/3)}]
This is a chunk from a video sequence. In the previous chunk, (the hand|the patient's LEFT hand|the patient's RIGHT hand) was holding a cylindrical block. Answer directly: 'Yes.' if (the hand|the patient's LEFT hand|the patient's RIGHT hand) puts down and releases the block; answer 'No.' otherwise.
\end{lstlisting}

\newpage
\section{Further Results for Primitives Identification}\label{app:further_results_prims_id}

Table~\ref{tab:sp_ablation} shows ablation results for prompting the primitive directly, as in Listing~\ref{lst:sp}. Table~\ref{tab:smc_ablation} shows results for using \textit{Decomposed Prompting} (see Listings~\ref{lst:dpm} and \ref{lst:dpg} for the prompts). We find that the performance of the two prompting methods to be comparable, and proceeded with \textit{Decomposed Prompting} in the main section to provide more granular analyses. 

The tables also compare different settings of the sampling rate ($f$) and number of frames per segment ($n$), sorted in decreasing segment duration. There is a trade-off between the three metrics: with decreasing segment duration, sequence predictions become more granular but also longer. This trend is reflected by a generally improving ES, and worsening AER and RCE. For \textit{Decomposed Prompting,} the best trade-off appears to occur with segment duration $0.5$ seconds. We chose the setting of $f=15$ and $n=8$ as a good balance of performance and compute speed for our model ablations.

% Both variants achieved comparable Edit Scores overall, indicating that neither approach was decisively superior in aggregate performance. However, Decomposed prompting was retained for follow-up analysis because Direct Prompting does not provide strict advantages, and Decomposed Prompting allows us to analyze errors in a more fine-grained way. Denser sampling (e.g., f=1, n=30) improved the Edit Score but doubled computation, while sparser settings underfit short actions. The (f=15, n=8) setup achieved a strong balance of high Edit Score (≈49.2) with low AER/RCE (≈0.64/0.65) at roughly half the compute cost of the denser variants and was adopted as the default for subsequent experiments.

\begin{table}[h!]
  \centering
  \caption{\textbf{Ablations for Direct Prompting using the prompt in Listing~\ref{lst:sp}.}}
\begin{tabular}{rlcccr}
\toprule $f$ & $n$ & ES $\uparrow$ & AER $\downarrow$ & RCE $\downarrow$ \\
\midrule
1 & 1  & 41.83 ± 1.01 & 0.65 ± 0.04 & 0.65 ± 0.04 \\
2 & 2  & 41.25 ± 1.07 & 0.64 ± 0.02 & 0.64 ± 0.03 \\
4 & 4  & 41.87 ± 1.19 & \cellcolor{blue!20}\textbf{0.62 ± 0.02} & \cellcolor{green!20}\textbf{0.60 ± 0.03} \\
8 & 8  & 40.40 ± 1.21 & 0.65 ± 0.03 & 0.64 ± 0.03 \\
15 & 15  & 41.86 ± 1.20 & \cellcolor{green!20}\textbf{0.61 ± 0.02} & \cellcolor{blue!20}\textbf{0.63 ± 0.03} \\
30 & 30  & 40.39 ± 1.24 & 0.63 ± 0.02 & 0.68 ± 0.02 \\
2 & 1  & 44.24 ± 1.01 & 0.81 ± 0.08 & 0.82 ± 0.08 \\
4 & 2  & 44.92 ± 1.00 & 0.82 ± 0.06 & 0.81 ± 0.06 \\
8 & 4  & \cellcolor{blue!20}\textbf{45.70 ± 1.02} & 0.69 ± 0.04 & 0.68 ± 0.04 \\
15 & 8  & \cellcolor{green!20}\textbf{46.34 ± 0.97} & 0.69 ± 0.05 & 0.71 ± 0.05 \\
30 & 15  & 43.73 ± 1.09 & 0.72 ± 0.05 & 0.77 ± 0.05 \\
4 & 1  & 38.56 ± 1.19 & 1.44 ± 0.14 & 1.45 ± 0.14 \\
8 & 2  & 37.81 ± 1.10 & 1.48 ± 0.13 & 1.46 ± 0.13 \\
15 & 4  & 40.25 ± 1.12 & 1.25 ± 0.11 & 1.28 ± 0.10 \\
30 & 8  & 41.39 ± 1.08 & 1.07 ± 0.07 & 1.08 ± 0.07 \\
8 & 1  & 28.44 ± 1.11 & 2.73 ± 0.24 & 2.73 ± 0.24 \\
\bottomrule
\end{tabular}
  \label{tab:sp_ablation}
\end{table}

\begin{table}[h!]
  \centering
     \caption{\textbf{Ablations for \textit{Decomposed Prompting} using the prompts in Listings~\ref{lst:dpm} and \ref{lst:dpg}.}}
    \begin{tabular}{rlcccr}
        \toprule
        $f$ & $n$ & ES $\uparrow$ & AER $\downarrow$ & RCE $\downarrow$ \\
        \midrule
            1 & 1 &  36.89 ± 1.63 & 0.68 ± 0.03 & 0.68 ± 0.03 \\
            2 & 2 &  34.52 ± 1.73 & 0.70 ± 0.03 & 0.70 ± 0.03 \\
            4 & 4 &  35.87 ± 1.70 & 0.67 ± 0.02 & 0.67 ± 0.02 \\
            8 & 8 &  37.45 ± 1.64 & 0.68 ± 0.04 & 0.67 ± 0.04 \\
            15 & 15 &  42.52 ± 1.47 & \cellcolor{green!20}\textbf{0.60 ± 0.02} & \cellcolor{green!20}\textbf{0.61 ± 0.03} \\
            30 & 30 &  42.71 ± 1.39 & \cellcolor{blue!20}\textbf{0.62 ± 0.04} & \cellcolor{blue!20}\textbf{0.64 ± 0.04} \\
            2 & 1 &  42.80 ± 1.64 & 0.73 ± 0.08 & 0.73 ± 0.08 \\
            4 & 2 &  40.68 ± 1.74 & 0.74 ± 0.08 & 0.75 ± 0.08 \\
            8 & 4 &  44.16 ± 1.75 & 0.71 ± 0.08 & 0.70 ± 0.08 \\
            15 & 8 &  46.69 ± 1.62 & 0.65 ± 0.06 & 0.65 ± 0.06 \\
            30 & 15 &  \cellcolor{blue!20}\textbf{49.21 ± 1.39} & 0.64 ± 0.08 & 0.65 ± 0.08 \\
            4 & 1 &  45.64 ± 1.68 & 0.93 ± 0.17 & 0.97 ± 0.17 \\
            8 & 2 &  44.57 ± 1.67 & 0.94 ± 0.17 & 1.00 ± 0.17 \\
            15 & 4 &  48.22 ± 1.69 & 0.86 ± 0.15 & 0.88 ± 0.15 \\
            30 & 8 &  \cellcolor{green!20}\textbf{50.76 ± 1.55} & 0.82 ± 0.13 & 0.84 ± 0.13 \\
            8 & 1 &  42.42 ± 1.64 & 1.45 ± 0.29 & 1.53 ± 0.28 \\
        \bottomrule
    \end{tabular}
  \label{tab:smc_ablation}
\end{table}

%In addition, in the motion analysis literature, videos are typically analyzed at $30+$ frames per second. Our initial ablations show that this level of granularity is expensive: since the visual tokens comprise the majority of the input tokens, inference time on a video increases roughly linearly with the number of frames processed (for a fixed number of frames per forward pass). Due to these considerations, we divided each video into $0.533$-s segments, corresponding to a sampling rate of $15$ frames per second with eight frames processed per forward pass.

\newpage
\section{Figure~\ref{fig:vlm_analysis}(d) Experimental Procedure}\label{app:left_v_right}

For Figure~\ref{fig:vlm_analysis}(d), we sourced videos from $5$ control subjects. These subjects performed the RTT task twice, once for either hand, filmed via two camera streams positioned at the front-left and front-right of the person (making $20$ videos in total). Crucially, the RTT task was designed so that only one hand performs the RTT task while the other hand remains still, with the first hand labeled. We chose segments where every frame is labeled with a moving primitive, i.e. one of \textit{transport}, \textit{reach}, or \textit{reposition}, leading to 1,101 segments of duration $0.533$ seconds. We then evaluated two methods for motion prediction on these segments: \textbf{Pred w/o Cropping}, where we prompted the model twice, once for each hand, asking if the hand was moving; and \textbf{Pred w/ Cropping}, where we tested if we could improve results by cropping around the desired hand and asking about the movements for the ``hand in the center." The prompts for both scenarios are listed below.

\begin{lstlisting}[caption={Cross-hand Prompting (Pred w/o Cropping) (1/2)}]
Focus on the patient's (LEFT/RIGHT) hand. Do not mention or consider the other hand in any way. Based on the movement and posture of the patient's (LEFT/RIGHT) hand, is the (LEFT/RIGHT) hand moving or moving an object? Answer 'Yes.' or 'No.' directly.
\end{lstlisting}

\begin{lstlisting}[caption={Cross-hand Prompting (Pred w/ Cropping---Even if Cropping Fails) (2/2)}]
Based on the movement and posture of the hand, is the hand in the center moving or moving an object? Answer 'Yes.' or 'No.' directly.
\end{lstlisting}

\section{More Details on the Fugl-Meyer Assessment Videos}
\label{app:fma_videos}

\begin{figure}[h!]
        \centering
        \includegraphics[width=1\linewidth]{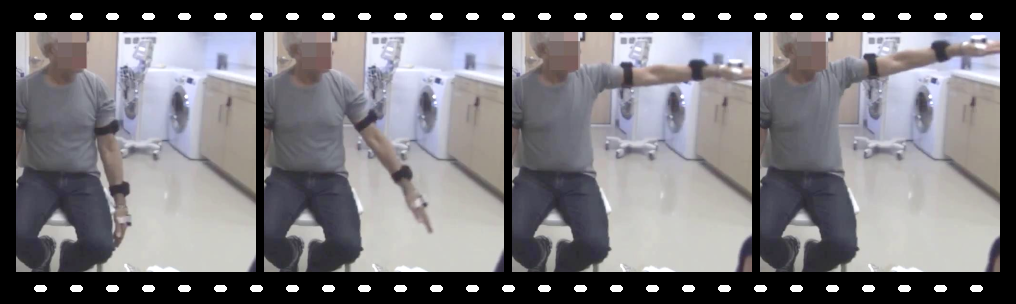}
        \caption{\textbf{Front view for the Fugl-Meyer assessment.}}
        \label{fig:fma_front_view}
\end{figure}

\begin{figure}[h!]
    \centering
    \includegraphics[width=\linewidth]{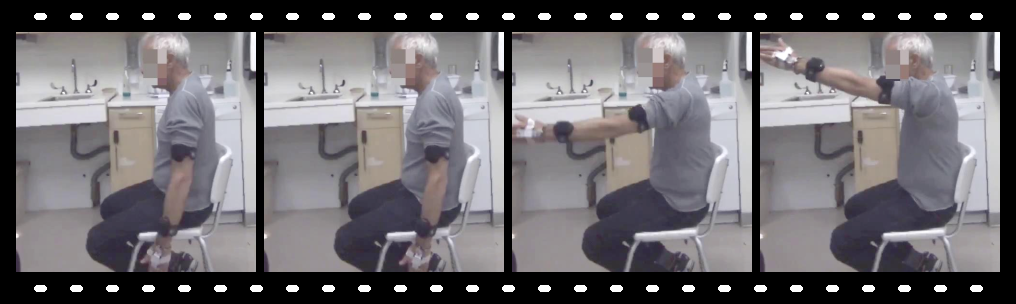}
    \caption{\textbf{Side view for the Fugl-Meyer assessment. (Different assessment item from Fig.~A\ref{fig:fma_front_view})}}
    \label{fig:fma_side_view}
\end{figure}

\begin{figure}[h!]
    \centering
    \includegraphics[width=\linewidth]{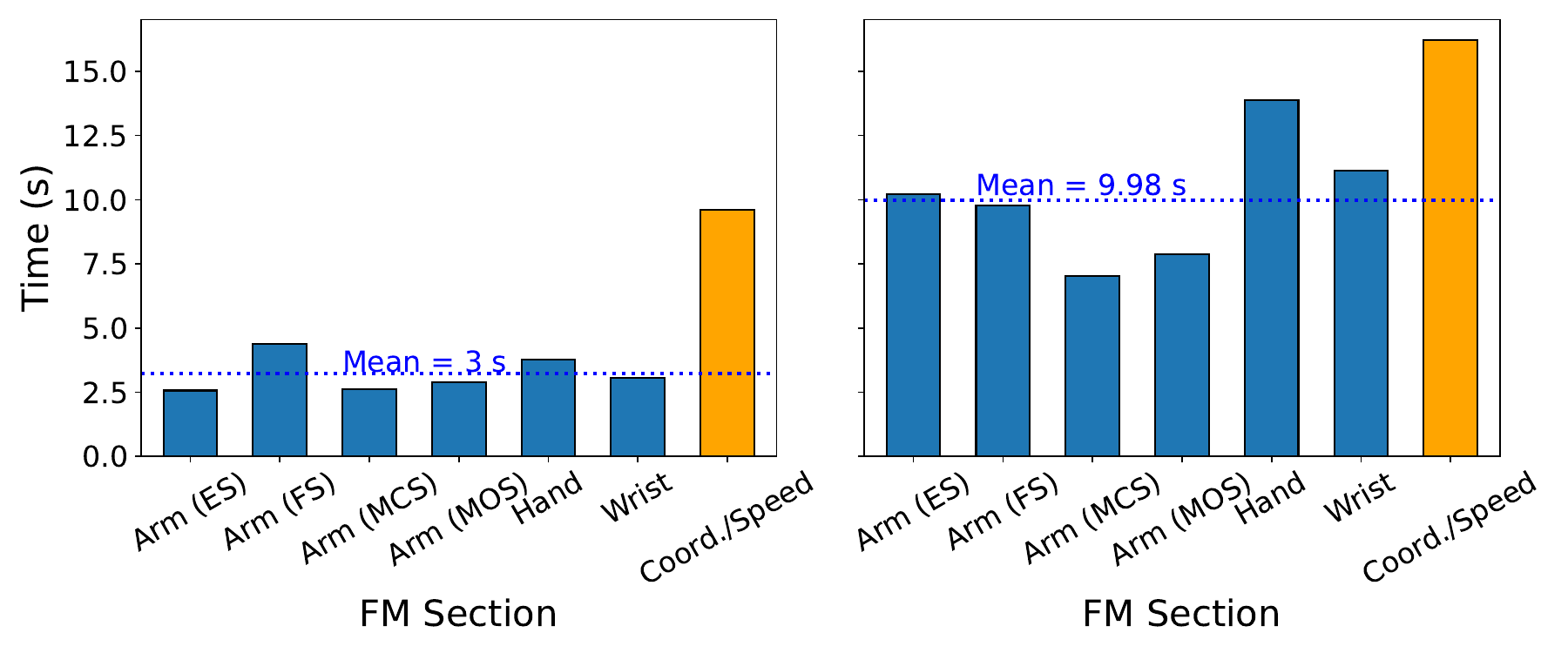}
    \caption{\textbf{Duration of Fugl-Meyer assessment videos by item section.} \textbf{Left:} Median video duration in seconds. \textbf{Right:} $95$-percentile video duration in seconds. The means are calculated using the videos in all sections except \textbf{Coord./Speed.}}
    \label{fig:fm_section_durations}
\end{figure}

Here, we illustrate the two views for the Fugl-Meyer assessment (Figs.~A\ref{fig:fma_front_view} and A\ref{fig:fma_side_view}), and justify the short video duration for videos not in the \textbf{Speed/Coord.} section (see Fig.~A\ref{fig:fm_section_durations}).

\section{Prompts for Impairment Quantification}\label{app:impairment_quant_prompts}

We include the prompts for \textbf{Rule-based Question-Answering} and \textbf{Chain-of-Thought} as supplementary CSV files: ``Supplementary-Data-FMA-QA.csv" and ``Supplementary-Data-FMA-CoT.csv", respectively. Table~\ref{tab:question_columns} describes each column in the files.

\begin{table}[h!]
\caption{\textbf{Description of the columns in the prompt CSV files}}
\centering
\small
\begin{tabular}{p{2.5cm} p{11cm}}
\toprule
\textbf{Column} & \textbf{Description} \\
\midrule
\texttt{qid} & Question identifier (integer index). \\[4pt]

\texttt{fm\_video} & Video identifier in the format \texttt{\{fm\_item\}\_\{side\}\_\{view\}}.  
\texttt{fm\_item} ranges from 3 to 33 and denotes the Fugl--Meyer (FM) assessment item.  
\texttt{side} is either \texttt{A} (affected) or \texttt{H} (healthy); most rows use \texttt{A}, but \texttt{H} is included for Coordination/Speed items to compare movement between sides.  
\texttt{view} indicates the camera view that best answers the question: \texttt{F} (front) or \texttt{S} (side). \\[6pt]

\texttt{question\_type} & Type of question, either \texttt{rate} (requires a numerical rating) or \texttt{binary} (yes/no response). \\[4pt]

\texttt{sampling} & Frame sampling method.  
\texttt{uniform}: sample 8 frames uniformly across the video.  
\texttt{dense}: segment the video into 0.267\,s chunks and sample 8 frames uniformly within each chunk. \\[6pt]

\texttt{binary\_no\_score} & For binary questions, if non-null, specifies the score to assign when the VLM predicts “no.” If null, the question chain continues. \\[4pt]

\texttt{binary\_yes\_score} & For binary questions, analogous to \texttt{binary\_no\_score}, but specifies the score to assign when the VLM predicts “yes.” \\[4pt]

\texttt{question} & The natural-language prompt given to the VLM (e.g.,  
“Assess the patient's maximal shoulder elevation at the ending position. Rate as follows: 0 (no elevation), 1 (partial elevation), or 2 (full elevation). Answer directly.”). \\
\bottomrule
\end{tabular}
\label{tab:question_columns}
\end{table}

\end{appendices}